\documentclass{article}

\usepackage[nonatbib, final]{nips_2017}

\usepackage[sort,numbers]{natbib}

\usepackage[utf8]{inputenc} 
\usepackage[T1]{fontenc}    
\usepackage[hidelinks]{hyperref}       
\usepackage{url}            
\usepackage{booktabs}       
\usepackage{amsfonts}       
\usepackage{nicefrac}       
\usepackage{microtype}      

\newcommand{\smartpaste}{\textsc{SmartPaste}\xspace}
\newcommand{\ourtitle}{\smartpaste: Learning to Adapt Source Code}

\usepackage{amssymb}       

\usepackage{multirow}
\usepackage{paralist}

\usepackage{algorithm}
\usepackage[noend]{algorithmic}

\usepackage{xspace}
\usepackage{color}
\usepackage{amsmath}
\usepackage{subcaption}
\usepackage{bm}

\newcommand{\cf}{\hbox{\emph{cf.}}\xspace}

\newcommand{\eg}{\hbox{\emph{e.g.}}\xspace}
\newcommand{\ie}{\hbox{\emph{i.e.}}\xspace}

\newcommand{\etc}{\hbox{\emph{etc.}}\xspace}

\newcommand{\CSharp}{C\nolinebreak\hspace{-.05em}\raisebox{.6ex}{\footnotesize\bf \#}}

\usepackage{ifthen}
\newboolean{showcomments}
\setboolean{showcomments}{true} 
\ifthenelse{\boolean{showcomments}}{
  \newcommand{\nbc}[3]{
    {\colorbox{#3}{\bfseries\sffamily\scriptsize\textcolor{white}{#1}}}%
    {\textcolor{#3}{\sf\small$\blacktriangleright$\textit{#2}$\blacktriangleleft$}}}
  \newcommand{\todo}[1]{\nbc{TODO}{#1}{blue}\xspace}
}{
  \newcommand{\nbc}[3]{}
  \newcommand{\todo}[1]{}
}

\usepackage{listings}
\newcommand{\id}[1]{\texttt{#1}}
\newenvironment{squishlist}
{
 \begin{list}{$\bullet$}
  { \setlength{\itemsep}{0pt}
     \setlength{\parsep}{3pt}
     \setlength{\topsep}{3pt}
     \setlength{\partopsep}{0pt}
     \setlength{\leftmargin}{1.5em}
     \setlength{\labelwidth}{1em}
     \setlength{\labelsep}{0.5em} } }
{  \end{list}  }

\DeclareMathOperator*{\argmax}{arg\,max}
\DeclareMathOperator*{\elmax}{el\,max}

\newcommand{\vect}[1]{\ensuremath{\mathbf{r}}\xspace}
\newcommand{\tok}{\ensuremath{t}\xspace}
\newcommand{\tokenSeq}{\ensuremath{\mathcal{T}}\xspace}
\newcommand{\var}{\ensuremath{v}\xspace}
\newcommand{\varSet}{\ensuremath{\mathbb{V}}\xspace}
\newcommand{\varType}[1]{\ensuremath{\tau}(#1)\xspace}
\newcommand{\allVarType}[1]{\ensuremath{\tau^*(#1)}\xspace}

\newcommand{\varsInScope}[1]{\ensuremath{\varSet_{#1}}\xspace}
\newcommand{\placeholderSet}{\ensuremath{\mathcal{P}}\xspace}
\newcommand{\assignment}{\ensuremath{\alpha}\xspace}
\newcommand{\lexPrevSym}{\ensuremath{\mathcal{U}}}
\newcommand{\lexPrev}[2]{\ensuremath{\lexPrevSym_{\mathsf{p}}(#1, #2)}}
\newcommand{\lexNext}[2]{\ensuremath{\lexPrevSym_{\mathsf{n}}(#1, #2)}}
\newcommand{\varUses}[1]{\ensuremath{\mathcal{U}}(#1)\xspace}
\newcommand{\dataFlowSym}{\ensuremath{\mathcal{D}}}
\newcommand{\dataFlowIn}[2]{\ensuremath{\dataFlowSym_{\mathsf{p}}(#1, #2)}}
\newcommand{\dataFlowOut}[2]{\ensuremath{\dataFlowSym_{\mathsf{n}}(#1, #2)}}

\newcommand{\typeEmbedSym}[0]{\ensuremath{\mathbf{r}}}
\newcommand{\typeEmbed}[1]{\ensuremath{\typeEmbedSym(#1)}}
\newcommand{\typeSetEmbed}[1]{\ensuremath{\typeEmbedSym^*(#1)}}
\newcommand{\tokenEmbedSym}[0]{\ensuremath{f}}
\newcommand{\tokenEmbed}[1]{\ensuremath{\tokenEmbedSym(#1)}}
\newcommand{\tokenSeqEmbedSym}[0]{\ensuremath{g}}
\newcommand{\contextSeqEmbedSym}[0]{\ensuremath{h}}

\newcommand{\localContextSym}[0]{\ensuremath{{\bm{c}}}}
\newcommand{\localContextRepr}[1]{\ensuremath{\localContextSym(#1)}}

\newcommand{\usageRepr}[2]{\ensuremath{\mathbf{u}(#1, #2)}}

\newcommand{\lexicalContextSym}{\ensuremath{\mathcal{U}^{\mathcal{L}}}}
\newcommand{\lexicalContextPrev}[3]{\ensuremath{\lexicalContextSym_{\mathsf{p}}(#1, #2, #3)}}
\newcommand{\lexicalContextNext}[3]{\ensuremath{\lexicalContextSym_{\mathsf{n}}(#1, #2, #3)}}
\newcommand{\lexicalContextRepr}[2]{\ensuremath{\mathbf{u}^{\mathcal{L}}(#1, #2)}}

\newcommand{\dataContextSym}{\ensuremath{\mathcal{U}^{\mathcal{D}}}}
\newcommand{\dataContextPrev}[3]{\ensuremath{\dataContextSym_{\mathsf{p}}(#1, #2, #3)}}
\newcommand{\dataContextNext}[3]{\ensuremath{\dataContextSym_{\mathsf{n}}(#1, #2, #3)}}
\newcommand{\dataContextRepr}[2]{\ensuremath{\mathbf{u}^{\mathcal{D}}(#1, #2)}}

\newcommand{\treeGRUPrev}[3]{\ensuremath{\mathbf{q}_{\mathsf{p}}}(#1, #2, #3)}
\newcommand{\treeGRUNext}[3]{\ensuremath{\mathbf{q}_{\mathsf{n}}}(#1, #2, #3)}

\newcommand{\weightsContext}{\ensuremath{\mathbf{W}_{\localContextSym}}}

\newcommand{\weightsDataRepr}{\ensuremath{\mathbf{W}_{\mathcal{D}}}}

\newcommand{\blankplaceholder}{\colorbox{black}{\textcolor{white}{\id{~~?~~}}}}
\newcommand{\namedplaceholder}[1]{\colorbox{blue}{\textcolor{white}{\id{~#1~}}}}
\newcommand{\placeholder}[1]{\colorbox{yellow}{\textcolor{black}{\id{#1}}}}

\usepackage{tikz}
\usetikzlibrary{arrows,decorations.pathmorphing,backgrounds,fit,positioning,automata,shapes,calc,shadows}
\tikzset{codegraph/.style={
                every node/.style={anchor=west, font=\tt,
                  text depth=0.5ex, text height=1.5ex, text centered, scale=.75,
                  inner sep=2pt},
                token/.append style={rectangle, draw=black, very thin},
                placeholder/.append style={rectangle, draw=red, thick, color=red, minimum width=5ex},
         }
}

\makeatletter
\newenvironment{btHighlight}[1][]
{\begingroup\tikzset{bt@Highlight@par/.style={#1}}\begin{lrbox}{\@tempboxa}}
{\end{lrbox}\bt@HL@box[bt@Highlight@par]{\@tempboxa}\endgroup}

\newcommand\btHL[1][]{%
  \begin{btHighlight}[#1]\bgroup\aftergroup\bt@HL@endenv%
}

\def\bt@HL@endenv{%
  \end{btHighlight}%
  \egroup
}
\newcommand{\bt@HL@box}[2][]{%
  \tikz[#1]{%
    \pgfpathrectangle{\pgfpoint{1pt}{0pt}}{\pgfpoint{\wd #2}{\ht #2}}%
    \pgfusepath{use as bounding box}%
    \node[anchor=base west, fill=blue!10,outer sep=0pt,inner xsep=1pt, rounded corners=1pt, inner ysep=0pt, minimum height=\ht\strutbox,#1]{\strut\usebox{#2}};
  }%
}
\makeatother

\newcommand{\hlreflarge}[1]{\begin{btHighlight}[fill=white!10,draw=green,solid,line width=.1pt]{$\lambda_{#1}$}\end{btHighlight}}
\newcommand{\hlref}[1]{$^\text{\tiny~\hlreflarge{#1}}$}

\newcommand{\hlplacehld}[2]{\begin{btHighlight}[fill=red!20,draw=red,solid,line width=.1pt]{#1$^{\lambda_{#2}}$}\end{btHighlight}}

\title{\ourtitle}

\author{
  Miltiadis~Allamanis\\
  Microsoft Research\\
  Cambridge, UK \\
  \texttt{t-mialla@microsoft.com} \\
  \And
  Marc~Brockschmidt\\
  Microsoft Research\\
  Cambridge, UK \\
  \texttt{mabrocks@microsoft.com}
}

\begin{document}
\maketitle

\begin{abstract}
Deep Neural Networks have been shown to succeed at a range of natural language
tasks such as machine translation and text summarization.
While tasks on source code (\ie, formal languages) have been considered
recently, most work in this area does not attempt to capitalize on the unique
opportunities offered by its known syntax and structure.
In this work, we introduce \smartpaste, a first task that requires to use such
information.
The task is a variant of the program repair problem that requires to adapt a
given (pasted) snippet of code to surrounding, existing source code.
As first solutions, we design a set of deep neural models that learn to
represent the context of each variable location and variable usage in a
data flow-sensitive way.
Our evaluation suggests that our models can learn to solve the \smartpaste task
in many cases, achieving 58.6\% accuracy, while learning meaningful
representation of variable usages.
\end{abstract}

\section{Introduction}
The advent of large repositories of source code as well as scalable machine
learning methods naturally leads to the idea of ``big code'', \ie, largely
unsupervised methods that support software engineers by generalizing from
existing source code.
Currently, existing machine learning models of source code capture
its shallow, textual structure, \eg
 as a sequence of tokens~\citep{hindle2012naturalness,allamanis2016convolutional},
 as parse trees~\citep{maddison2014structured,bielik2016phog}, or
 as a flat dependency networks of variables~\citep{raychev2015predicting}.
Such models miss out on the opportunity to capitalize on the rich and
well-defined semantics of source code.
In this work, we take a step to alleviate this by taking advantage of two
additional elements of source code: data flow and execution paths.
Our key insight is that exposing these semantics explicitly as input to a
machine learning model lessens the requirements on amounts of training data,
model capacity and training regime and allows us to solve tasks that are beyond
the current state of the art.

To show how this information can be used, we introduce the \smartpaste structured prediction task,
in which a larger, existing piece of source code is extended by a new snippet
of code and the variables used in the pasted code need to be aligned with
the variables used in the context. This task can be seen as a constrained
code synthesis task and simultaneously as a useful machine learning-based software
engineering tool. To achieve high accuracy on \smartpaste, we need to learn
representations of program semantics. First an
approximation of the semantic role of a variable (\eg, ``is it a counter?'', ``is
it a filename?'') needs to be learned. Second, an approximation of variable usage
semantics (\eg, ``a filename is needed here'') is required.
``Filling the blank element(s)''  is related to methods for learning distributed
representations of natural language words, such as Word2Vec~\citep{mikolov2013distributed}
and GLoVe~\citep{pennington2014glove}.
However, in our setting, we can learn from a much richer structure, such as
data flow information.
Thus, \smartpaste can be seen as a first step towards learning distributed
representations of variable usages in an unsupervised manner.
We expect such representations to be valuable in a wide range of tasks,
such as code completion (``this is the variable you are looking for''), bug
finding (``this is \emph{not} the variable you are looking for''), and summarization
(``such variables are usually called \id{filePath}'').

To summarize, our contributions are:
\begin{inparaenum}[(i)]
\item We define the \smartpaste task as a challenge for machine
 learning modeling of source code, that requires to learn (some) semantics of
 programs (\cf \autoref{sec:task}).
\item We present five models for solving the \smartpaste task by modeling it as
  a probability distribution over graph structures which represent code's data
  flow (\cf \autoref{sec:models}).
\item We evaluate our models on a large dataset of 4.8 million lines of
 real-world source code, showing that our best model achieves accuracy
 of 58.6\% in the \smartpaste task while learning useful
 vector representations of variables and their usages (\cf \autoref{sec:evaluation}).
\end{inparaenum}


\section{The \smartpaste Task}
\label{sec:task}
\begin{figure}
  \centering
\begin{subfigure}[b]{.54\textwidth}
\begin{minipage}{\textwidth}
\begin{lstlisting}[xleftmargin=0cm]
int SumPositive(int[] arr(*\hlref{0}*), int lim(*\hlref{1}*)) {
    int sum(*\hlref{2}*)=0;
   (*\hspace{-.8mm}\tikz[remember picture] \node [] (regStart) {};*)for(int i(*\hlref{3}*)=0; (*\hlplacehld{i}{4}*)<(*\hlplacehld{lim}{5}*); (*\hlplacehld{i}{6}*)++)
       if ((*\hlplacehld{arr}{7}*)[(*\hlplacehld{i}{8}*)]>0) (*\hlplacehld{sum}{9}*)+=(*\hlplacehld{arr}{10}*)[(*\hlplacehld{i}{11}*)];              (*\tikz[remember picture] \node [] (regEnd) {};*)
    return sum(*\hlref{12}*);
}
\end{lstlisting}
\begin{tikzpicture}[remember picture, overlay]
    \draw[draw=black, dashed, fill=green!40, rounded corners,fill opacity=.1] ($(regStart.north west) + (.1, .1)$) rectangle ($(regEnd.south) - (.1, .15)$);
\end{tikzpicture}
\end{minipage}
\caption{Example source code, with pasted snippet shaded in green.
  Tokens corresponding to variables marked by $\lambda_i$.
  The red boxes are the placeholders whose variable needs to be inferred 
  in the \smartpaste{} task (ground truth variable
  name shown for convenience).}
\end{subfigure}
\hfill
\begin{subfigure}[b]{.4\textwidth}
\begin{minipage}{\textwidth}
    \tikzset{placeholdergraph/.style={
                    every node/.style={anchor=west, font=\tt,
                    text depth=0.5ex, text height=1.5ex, text centered, scale=.75,
                    inner sep=2pt},
                    knownplaceholder/.append style={rectangle, draw=red, color=black},
                    placeholder/.append style={rectangle, fill=red!10,draw=red,solid,line width=.5pt},
            }
    }

  \begin{tikzpicture}[placeholdergraph, node distance=6pt and 5pt]
    \node[knownplaceholder] (l0) at (0,0) {$\lambda_0$:arr};
    \node[knownplaceholder] (l1) [right= of l0] {$\lambda_1$:lim};
    \node[knownplaceholder] (l3) [right= of l1] {$\lambda_3$:i};
    \node[knownplaceholder] (l2) [right= of l3] {$\lambda_2$:sum};
    
    \node[placeholder] (l4) [below=of l3] {$\lambda_4$:i};
    \node[placeholder] (l5) [below=of l1] {$\lambda_5$:lim};

    \node[placeholder] (l7) [below=of l0] {$\lambda_7$:arr};
    \node[placeholder] (l8) [below=of l5] {$\lambda_8$:i};

    \node[placeholder] (l11) [below=of l8] {$\lambda_{11}$:i};
    \node[placeholder] (l10) [left=of l11] {$\lambda_{10}$:arr};
    \node[placeholder] (l9) [right=of l11] {$\lambda_{9}$:sum};    
    \node[placeholder] (l6) [right=of l9] {$\lambda_6$:i};

    \node[knownplaceholder] (l12) [right=of l6] {$\lambda_{12}$:sum};

    \draw[thick, dashed, blue]
      (l2.south east) edge[->, bend left=30] (l12.north)
    ;

     \draw[thick, dotted, red]
      (l0.south) edge[->] (l7.north)
      (l1.south) edge[->] (l5.north)
      (l2.south) edge[->, bend right=5] (l9.north)   
      (l7.south) edge[->, bend right=5] (l10.north)  
      (l9.south) edge[->, bend right=40] (l12.south)
      (l3.south) edge[->] (l4.north)
      (l4.south) edge[->, bend left=13] (l8.east)
      (l8.south) edge[->] (l11.north)
      (l11.south) edge[->, bend right=40] (l6.south)
      (l8.east) edge[->, bend left=10] (l6.north)
    ;

  \end{tikzpicture}
\end{minipage}
\caption{Dataflow diagram for variables in example, using ground truth
  placeholder choices.
  Dotted red edges show dataflow that depends on placeholder allocations.
  Dashed blue edge is dataflow independent of choices.\label{fig:smartpastegraph}}
\end{subfigure}
  \caption{\label{fig:CodeGraphExample} The snippet (shaded box, left) was
    pasted into the existing code.
    Our task is to assign variables to each placeholder (red boxes).
    This requires inferring the flow of data between placeholders (right).}\vspace{-1em}
\end{figure}
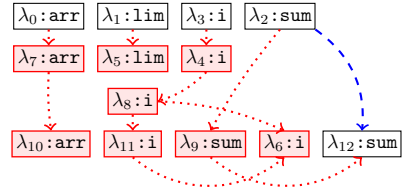

We consider a task beyond standard source code completion in which we
want to insert a snippet of code into an existing program and adapt variable
identifiers in the snippet to fit the target program (\autoref{fig:CodeGraphExample}).
This is a common scenario in software development~\citep{amann2016study}, when
developers copy a piece of code from a website (\eg StackOverflow) or
from an existing project into a new context. Furthermore,
pasting code is a common source of software bugs \citep{ray2013detecting}, with
more than 40\% of Linux porting bugs caused by the inconsistent renaming
of identifiers.

While similar to standard code completion, this task differs in a number of
important aspects. First, only variable identifiers need to be filled in, whereas many code
completion systems focus on a broader task (\eg predicting every next token
in code).
Second, several identifiers need to be filled in at the same time and thus all
choices need to be made synchronously, reflecting interdependencies. This amounts
to the structured prediction problem of inferring a graph structure (\cf \autoref{fig:smartpastegraph}).

\paragraph{Task Description.}
We view a source code file as a sequence of tokens $\tok_0 \dots \tok_N = \tokenSeq$.
The source code contains a set of variables $\var_0, \var_1 \dots \in
\varSet \subseteq \tokenSeq$.
To simplify the presentation, we assume that the source snippet to be pasted has
already been inserted at the target location, and all identifiers in it have
been replaced by a set \placeholderSet{} of fresh placeholder
identifiers (see \autoref{fig:CodeGraphExample} for an example).

Thus, our input is a sequence of tokens $\tok_0 \dots \tok_N$ with
$\{\tok_{\lambda_1}, \dots, \tok_{\lambda_K}\} = \placeholderSet$, and our aim is
to find the ``correct'' assignment $\assignment: \placeholderSet \to \varSet$ of
variables to placeholders.
For training and evaluation purposes, a correct solution is one that simply
matches the ground truth, but note that in practice, several possible
assignments could be considered correct.


\section{Models}
\label{sec:models}
In the following, we discuss a sequence of models designed for the \smartpaste
task, integrating more and more known semantics of the underlying programming
language.
All models share the concepts of
 a \emph{context representation} $\localContextRepr{\tok}$ of a token $\tok$ and
 a \emph{usage representation} $\usageRepr{\tok}{\var}$ of the usage of a variable
  $\var$ at token $\tok$.
The models differ in the definitions $\localContextRepr{\tok}$ and
$\usageRepr{\tok}{\var}$, but all finally try to maximize the inner product of
$\localContextRepr{\tok}$ and $\usageRepr{\tok}{\var}$ for the correct
variable assignment $\var$ at $\tok$.

\paragraph{Notation}
We use $\varsInScope{\tok} \subset \varSet$ to refer to the set of all
variables in scope at the location of $\tok$, \ie, those variables that can be
legally used at $\tok$.
Furthermore, we use $\lexPrev{\tok}{\var} \in \tokenSeq \cup \{ \bot \}$ to
denote the last occurrence of variable $\var$ before $\tok$ in $\tokenSeq$
(resp. $\lexNext{\tok}{\var}$ for the next occurrence), where $\bot$ is used when
no previous (resp.\@ next) such token exists.
To denote all uses of a variable $\var \in \varSet$, we use $\varUses{\var}
\subset \tokenSeq$.
To capture the flow of data through a program, we furthermore introduce the
notation $\dataFlowIn{\tok}{\var} \subseteq \tokenSeq$, which denotes the set of tokens
at which $\var$ was possibly last used in an execution of the program (\ie
either read from or written to).
Similarly, $\dataFlowOut{\tok}{\var}$ denotes the tokens at which $\var$
is next used.
Note that $\dataFlowIn{\tok}{\var}$ (resp. $\dataFlowOut{\tok}{\var}$) is a set of
tokens and extends the notion of $\lexPrev{\tok}{\var}$ (resp. $\lexNext{\tok}{\var}$) which
refers to a single token.
Furthermore $\dataFlowIn{\tok}{\var}$ may include tokens appearing \emph{after}
\tok (resp.\@ $\dataFlowOut{\tok}{\var}$ may include tokens appearing
\emph{before} \tok) in the case of loops, as it happens for variable \id{i} in
$\lambda_{11}$ and $\lambda_{6}$ in \autoref{fig:CodeGraphExample}. 
$\dataFlowIn{\tok}{\var}$ and $\dataFlowOut{\tok}{\var}$ for the snippet in 
\autoref{fig:CodeGraphExample} are depicted in \autoref{fig:computation}.

\paragraph{Leveraging Variable Type Information}
We assume a statically typed language and that the source code can be compiled,
and thus each variable has a (known) type $\varType{\var}$.
To use it, we define a learnable embedding function $\typeEmbed{\tau}$ for known
types and additionally define an ``\textsc{UnkType}'' for all
unknown/unrepresented types.
We also leverage the rich type hierarchy that is available in many
object-oriented languages.
For this, we map a variable's type $\varType{\var}$ to the set of its
supertypes, \ie
$\allVarType{\var}=\{\tau : \varType{\var} \text{ implements type } \tau \} \cup \{ \varType{\var} \}$.
We then compute the type representation $\typeSetEmbed{\var}$ of a variable
$\var$ as the element-wise maximum of $\{ \typeEmbed{\tau} : \tau \in \allVarType{\var}\}$.
We chose the maximum here, as it is a natural pooling operation for representing
partial ordering relations (such as type lattices).
Using all types in $\allVarType{\var}$ allows us to generalize to unseen types
that implement common supertypes or interfaces.
For example, \id{List<K>} has multiple concrete types  (\eg \id{List<int>},
\id{List<string>}).
Nevertheless, these types implement a common interface (\id{IList}) and share
common characteristics.
During training, we randomly select a non-empty subset of $\allVarType{\var}$
which ensures training of all known types in the lattice. This acts both like
a dropout mechanism and allows us to learn a good representation for types
that only have a single known subtype in the training data.

\paragraph{Context Representations}
To fill in placeholders, we need to be able to learn how they are used.
Intuitively, usage is defined by the source code surrounding the
placeholder, as it describes what operations are performed on it.
Consequently, we define the notion of a \emph{context} of a token
$\tok_k$ as the sequences of $C$ tokens before and after $\tok_k$ (we use $C=3$).
We use a learnable function \tokenEmbedSym{} that embeds each token $\tok$
separately into a vector $\tokenEmbed{\tok}$ and finally compute the \emph{context
representation} \localContextRepr{\tok} using two learnable functions 
$\tokenSeqEmbedSym{}^p$ and $\tokenSeqEmbedSym{}^n$ to
combine the token representations as follows.
\begin{equation*}
  \localContextRepr{\tok_k} =
  \mathbf{W}_{\localContextSym} \cdot
  \left[\tokenSeqEmbedSym^p\left(\tokenEmbed{\tok_{k-C}}, \ldots, \tokenEmbed{\tok_{k-1}}\right),
        \tokenSeqEmbedSym^n\left(\tokenEmbed{\tok_{k+1}}, \ldots, \tokenEmbed{\tok_{k+C}}\right)
  \right]
\end{equation*}
Here, \weightsContext{} is a simple (unbiased) linear layer.
Note that we process the representation of preceding and succeeding tokens
separately, as the semantics of tokens strongly depends on
their position relative to $\tok$.
In this work, we experiment with a log-bilinear model~\citep{mnih2012fast} and a
GRU~\citep{cho2014properties} for $\tokenSeqEmbedSym$.
Our embedding function \tokenEmbed{\tok} integrates type information as follows.
If $\tok$ is a variable (\ie $\tok \in \var$) it assigns $\typeSetEmbed{\var}$ to $\tokenEmbed{\tok}$.
For each non-variable tokens $\tok$, it returns a learned embedding $\vect{r}_t$.

\paragraph{Usage Representations}
We learn a vector representation \usageRepr{\tok}{\var} as an approximation of
the semantics of a variable $\var$ at position $\tok$ by considering how it has
been used \emph{before and after} $\tok$.
Here, we consider two possible choices of representing usages, namely the
\emph{lexical usage representation} and the \emph{data flow usage
  representation} of a variable.

First, we view source code as a simple sequence of tokens.
We define the lexical usage representation \lexicalContextRepr{\tok}{\var} of
a variable $\var$ at placeholder $\tok$ using up to $L$ (fixed to 14 during
training\footnote{We set $L=14$ to capture the 98th percentile of our training
data and also allow efficient batching with padding instead of choosing the
maximum $L$ in the data.}) usages of $\var$ around $\tok$ in lexical order.
For this, we use our learnable context representation $\localContextSym$, and
define a sequence of preceding (resp.\@ succeeding) usages of a variable
recursively as follows.
\begin{align*}
\lexicalContextPrev{L}{\tok}{\var} =
    \begin{cases}
      \lexicalContextPrev{L-1}{\tok'}{\var} \circ \localContextRepr{\tok'}
        & \text{if } L > 0 \land \tok' = \lexPrev{\tok}{\var} \neq \bot\\
      \epsilon & \text{otherwise}
    \end{cases}\\
\lexicalContextNext{L}{\tok}{\var} =
    \begin{cases}
      \localContextRepr{\tok'} \circ \lexicalContextNext{L-1}{\tok'}{\var}
        & \text{if } L > 0 \land \tok' = \lexNext{\tok}{\var} \neq \bot\\
      \epsilon & \text{otherwise}
    \end{cases}
\end{align*}
Here, $\circ$ is sequence composition and $\epsilon$ is the empty sequence.
Then, we can define
$\lexicalContextRepr{\tok}{\var} =
 \contextSeqEmbedSym(\lexicalContextPrev{L}{\tok}{\var},
                     \lexicalContextNext{L}{\tok}{\var})$,
\ie the combination of the representations of the surrounding contexts.
We will discuss two choices of $\contextSeqEmbedSym$ below, namely averaging and
a RNN-based model. Note that \localContextRepr{\tok} is \emph{not} included in
either \lexicalContextPrev{L}{\tok}{\var} or \lexicalContextNext{L}{\tok}{\var}

Our second method for computing \usageRepr{\tok}{\var} takes the flow of data
into account.
Instead of using lexically preceding (resp.\@ succeeding) contexts, we consider
the data flow relation to identify relevant contexts.
Unlike before, there may be several predecessors (resp.\@ successors) of a
variable use in the data flow relationship, \eg to reflect a conditional operation
on a variable.
Thus, we define a \emph{tree} of $D$ preceding contexts
\dataContextPrev{D}{\tok}{\var}, re-using our context representation
$\localContextSym$, as a limited unrolling\footnote{$D=15$ during training to 
covers the 98th percentile in our training data and allows us to batch.} of the data flow graph as follows.
\begin{equation*}
  \dataContextPrev{D}{\tok}{\var} =
    \begin{cases}
      \{ (\localContextRepr{\tok'_0}, \dataContextPrev{D-1}{\tok'_0}{\var}),
           \ldots,
           (\localContextRepr{\tok'_d}, \dataContextPrev{D-1}{\tok'_d}{\var}) \}
        & \text{if } D > 0 \land \dataFlowIn{\tok}{\var} = \{ \tok'_0, \ldots, \tok'_d \}\\
      \emptyset
        & \text{otherwise}
    \end{cases}
\end{equation*}
The tree of $D$ succeeding contexts \dataContextNext{D}{\tok}{\var} is defined
analogously.
For example, \autoref{fig:computation} shows
 $\dataContextPrev{2}{\lambda_8}{\cdot}$ and
 $\dataContextNext{2}{\lambda_8}{\cdot}$
for all variables in scope at $\lambda_8$ of \autoref{fig:CodeGraphExample}.
We then compute a representation for the trees using a recursive
neural network, whose results are then combined with an unbiased linear layer to obtain
$\dataContextRepr{\tok}{\var}$. Again, \localContextRepr{\tok} is \emph{not} in
either \dataContextPrev{D}{\tok}{\var} or \dataContextNext{D}{\tok}{\var}.

Note that in this way of computing the context, lexically distant variables uses
(\eg before a long conditional block or a loop) can be taken into account when
computing a representation.

\subsection{Learning to Paste}
Using the context representation \localContextRepr{\tok} of the placeholder
$\tok$ and the usage representations \usageRepr{\tok}{\var} of all variables
$\var \in \varsInScope{}$ we can now formulate the probability of a single
placeholder $\tok$ being filled by a variable $\var$ as the inner product of the
two vectors:
\begin{equation*}
  p(\tokenSeq[\text{replace }\tok \text{ by } \var]) \propto
    \left(\localContextRepr{\tok}\right)^T \cdot \usageRepr{\tok}{\var}
\end{equation*}
When considering more than one placeholder, we aim to find an assignment $\assignment: \placeholderSet \to \varSet$ 
such that it maximizes the probability of the code obtained by replacing all
placeholders $\tok \in \placeholderSet$ according to \assignment at the same time, \ie,
\begin{align} \label{eq:testobj}
  \argmax_{\assignment}
       p(\tokenSeq[\text{replace all }\tok \in \placeholderSet \text{ by } \assignment(\tok)]).
\end{align}

As in all structured prediction models, training the model directly on \autoref{eq:testobj} is computationally
intractable because the normalization constant requires to compute
exponentially (up to $|\varSet|^{|\placeholderSet|}$) many different assignments.
Thus, during training, we choose to train on a single usage, \ie
$\max_{\boldsymbol\theta} p(\tokenSeq[\text{replace }\tok
\text{ by } \assignment(\tok) \text{ and all others are fixed to ground truth}])$ where $\boldsymbol\theta$
are all the parameters of the trained model.
However, this objective is still computationally expensive since it requires to
compute \usageRepr{\tok}{\var} for all $\var$ of the variably-sized $|\varSet|$ per placeholder.
To circumvent this problem and allow efficient batching,
we approximate the normalization constant by using all
variables in the current minibatch and train using maximum likelihood.

\begin{figure}[t]
 \tikzset{computegraph/.style={
                    every node/.style={anchor=west, font=\tt,
                    text depth=0.5ex, text height=1.5ex, text centered, scale=.75,
                    inner sep=2pt},
                    knownplaceholder/.append style={rectangle, draw=red, color=black, minimum width=5ex},
                    placeholder/.append style={rectangle, draw=red,solid,line width=.5pt},
                    usagecx/.append style={rectangle, draw=blue, fill=blue!10, solid,line width=.5pt},
        }
}
\centering
\begin{subfigure}[b]{.45\textwidth}
\begin{minipage}{\textwidth}\centering
  \begin{tikzpicture}[computegraph]
    \node[knownplaceholder] (l0) at (0,0) {$\lambda_0$};
    \node[knownplaceholder] (l1) [right of=l0] {$\lambda_1$};
    \node[knownplaceholder] (l2) [right of=l1] {$\lambda_2$};
    \node[knownplaceholder] (l3) [right of=l2] {$\lambda_3$};
    
    \node[placeholder] (l4) [below of=l3] {$\lambda_4$};
    \node[placeholder] (l5) [below of=l1] {$\lambda_5$};

    \node[placeholder] (l7) [below of=l0] {$\lambda_7$};

    \node[usagecx] (beforearr) [below of=l7] {arr};
    
     \node[usagecx] (beforelim) [right of=beforearr] {lim};

     \node[usagecx] (beforesum) [right of=beforelim] {sum};    
    
    \node[usagecx] (beforei) [right of=beforesum] {i};    

    \node (beforetitle) [right of=beforei] {$\dataFlowIn{\lambda_8}{\var_i}$};
 
     \draw[thick, blue, dotted]
      (l0.south) edge[->] (l7.north)
      (l7) edge[->] (beforearr)
      (l5) edge[->] (beforelim)
      (l4) edge[->] (beforei)
      (l2) edge[->] (beforesum)
      (l1.south) edge[->] (l5.north)  
      (l3.south) edge[->] (l4.north)
    ;

    \begin{scope}[on background layer]
      \draw[color=blue!30, dashed, rounded corners]
        ($(beforearr.north west) + (-0.13, 0.13)$)
        rectangle
        ($(beforetitle.south east) - (-0.13, 0.13)$);
    \end{scope}
  \end{tikzpicture}
\end{minipage}
\caption{$\dataFlowIn{\lambda_8}{\var_i}$ for in-scope variables at $\lambda_8$}
\end{subfigure}
\hfill
\begin{subfigure}[b]{.45\textwidth}
\begin{minipage}{\textwidth}\centering
  \begin{tikzpicture}[computegraph]
    \node[usagecx] (afterarr) at (0,0)  {arr};
    \node[usagecx] (afterlim) [right of=afterarr] {lim};    
    \node[usagecx] (aftersum) [right of=afterlim] {sum};
    \node[usagecx] (afteri) [right of=aftersum] {i};
    \node (aftertitle) [right of=afteri] {$\dataFlowOut{\lambda_8}{\var_i}$};

    \node[knownplaceholder, dashed] (empty) [below of=afterlim] {$\epsilon$};
    \node[placeholder] (l10) [left of=empty] {$\lambda_{10}$};
    \node[placeholder] (l9) [right of=empty] {$\lambda_{9}$};    
    \node[placeholder] (l11) [right of=l9] {$\lambda_{11}$};
    \node[placeholder] (l6) [right of=l11] {$\lambda_6$};

    \node[knownplaceholder] (l12) [right of=l6] {$\lambda_{12}$};

     \draw[thick, blue, dotted] 
      (l12.south) edge[->, bend left=50] (l9.south)
      (l6.south) edge[->, bend left=50] (l11.south)
      (l10) edge[->] (afterarr)
      (l11) edge[->] (afteri)
      (l6) edge[->] (afteri)
      (l9) edge[->] (aftersum)
     (empty) edge[->] (afterlim)
     (empty) edge[->] (afterarr) 
    ;

    \begin{scope}[on background layer]
      \draw[color=blue!30, dashed, rounded corners]
        ($(afterarr.north west) + (-0.13, 0.13)$)
        rectangle
        ($(aftertitle.south east) - (-0.13, 0.13)$);
    \end{scope}
  \end{tikzpicture}
\end{minipage}
\caption{$\dataFlowOut{\lambda_8}{\var_i}$ for in-scope variables at $\lambda_8$}
\end{subfigure}
 \caption{\label{fig:computation}$\dataFlowIn{\lambda_8}{\var}$ and $\dataFlowOut{\lambda_8}{\var}$
 for the code in \autoref{fig:CodeGraphExample}. For each in-scope candidate $\var \in \varsInScope{\lambda_8}$, a representation
 is computed using the usage context of that variable before and after that placeholder. Then, the variable
 $\var^* = \argmax_{\var} \left(\localContextRepr{\lambda_8}\right)^T \cdot \usageRepr{\lambda_8}{\var}$ is selected
 for the placeholder. Arrows show the
 dataflow dependencies of each variable at $\lambda_8$ if that variable was to
 be used at this placeholder.}
\end{figure}

At test time, we need to fill in several placeholders in a given snippet of
inserted code.
To solve this structured prediction problem, we resort to iterative conditional
modes (ICM), where starting from a random allocation \assignment,
iteratively for each placeholder $\tok$, we pick the variable 
$v^* = \argmax_{\var \in \varSet_{\tok}} p(\tokenSeq[\text{replace }\tok \text{ by } \var])$
until the assignment map \assignment converges or we reach a maximum number of
iterations.
To recover from local optima, we restart the search a few times; selecting the
allocation with the highest probability.
Note that \lexPrev{\tok}{\var}, \lexNext{\tok}{\var}, 
$\dataFlowIn{\tok}{\var}$, $\dataFlowOut{\tok}{\var}$ 
and thus $\usageRepr{\tok}{\var}$ change during ICM, as the
underlying source code is updated to reflect the last chosen assignment
$\assignment$.

\paragraph{Model Zoo}
\newcommand{\localmodel}{\textsc{Loc}\xspace}
\newcommand{\lblmodel}{\ensuremath{\textsc{Avg}\mathcal{G}}\xspace}
\newcommand{\grumodel}{\ensuremath{\textsc{Gru}\mathcal{G}}\xspace}
\newcommand{\grudfmodel}{\ensuremath{\textsc{Gru}\mathcal{D}}\xspace}
\newcommand{\hybridmodel}{\ensuremath{\textsc{h}\mathcal{D}}\xspace}
We evaluate 5 different models in this work, based on different choices for
the implementation of $\usageRepr{\tok}{\var}$ and $\localContextRepr{\tok}$.

\begin{squishlist}
  \item \localmodel is a baseline using only local type information, \ie
    $\usageRepr{\tok}{\var}=\typeSetEmbed{\var}$.
 \item \lblmodel averages over the (variable length) context representations of
   the lexical context,
   \ie
   \begin{equation*}
     \usageRepr{\tok}{\var} =
         \typeSetEmbed{\var}
         + \frac{1}
                {\left|\lexicalContextPrev{L}{\tok}{\var}\right| + \left|\lexicalContextNext{L}{\tok}{\var}\right|}
           \left( 
             \sum_i (\lexicalContextPrev{L}{\tok}{\var})_i
             +
             \sum_i (\lexicalContextNext{L}{\tok}{\var})_i
           \right).
   \end{equation*}
 \item \grumodel uses a combination of the outputs of two GRUs to
   process the representations of the lexical context,
   \ie
   \begin{equation*}
     \usageRepr{\tok}{\var} =
        \mathbf{W}_{gru}
          \cdot
        \left[
          \textsc{RNN}_{\textsc{GRU}}^p(\lexicalContextPrev{L}{\tok}{\var}),
          \textsc{RNN}_{\textsc{GRU}}^n(\lexicalContextNext{L}{\tok}{\var})
        \right],
   \end{equation*}
   where $\mathbf{W}_{gru}$ is a learned (unbiased) linear layer. Note that the
   two RNNs have different learned parameters.
   The initial state of the $\textsc{RNN}_{\textsc{GRU}}$ is set to $\typeSetEmbed{\var}$.
 \item \grudfmodel uses two TreeGRU models (akin to TreeLSTM of \citet{tai2015improved},
   but using a GRU cell) over the tree structures \dataContextPrev{D}{\tok}{\var} and
   \dataContextNext{D}{\tok}{\var}, where we pool the representations
   computed for child nodes using an element-wise maximum operation $\elmax$.
   The state of leafs of the data flow tree are again initialized with the type
   embedding of $\var$, and thus, we have
   \begin{equation*}
     \treeGRUPrev{D}{\tok}{\var} =
       \begin{cases}
         \elmax_{\tok' \in \dataFlowIn{\tok}{\var}}(\textsc{GRU}(
                          \localContextRepr{\tok'},
                        \treeGRUPrev{D-1}{\tok'}{\var}))
            & \text{if } D > 0 \land \dataFlowIn{\tok}{\var}\neq\emptyset\\
         \typeSetEmbed{\var}
            & \text{otherwise}.
       \end{cases}
   \end{equation*}
   Analogously, we define \treeGRUNext{D}{\tok}{\var} and combine them to obtain
   \begin{equation*}
     \usageRepr{\tok}{\var} =
        \weightsDataRepr
          \cdot
        \left[
          \treeGRUPrev{D}{\tok}{\var},
          \treeGRUNext{D}{\tok}{\var}
        \right],
   \end{equation*}
   where \weightsDataRepr{} is a learned (unbiased) linear layer.
 \item \hybridmodel is a hybrid between \lblmodel and \grudfmodel, which uses
   another linear layer to combine their usage representations into a single
   representation of the correct dimensionality.
\end{squishlist}


\section{Evaluation}
\label{sec:evaluation}
\paragraph{Dataset}
We collected a dataset for the \smartpaste task from open source \CSharp{}
projects on GitHub.
To select projects, we picked the top-starred (non-fork) projects in GitHub.
We then filtered out projects that we could not (easily) compile in full using
Roslyn\footnote{\url{http://roslyn.io}}, as we require a compilation to extract
precise type information for the code (including those types present in external
libraries).
Our final dataset contains 27 projects from a diverse set of domains (compilers,
databases, \ldots) with about 4.8 million non-empty lines of code.
A full table is shown in \autoref{app:dataset}.

We then created \smartpaste examples by selecting snippets of up to 80 syntax
tokens (in practice, this means snippets are about 10 statements long)
from the source files of a project that are either children of a single AST node 
(\eg a block or a \id{for} loop) or are a contiguous sequence of statements. We then replace all
variables in the pasted snippet by placeholders.
The task is then to infer the variables that were replaced by placeholders.

\newcommand{\unseenTest}{\textsc{UnseenProjTest}\xspace}
\newcommand{\seenTest}{\textsc{SeenProjTest}\xspace}
From our dataset, we selected two projects as our validation set.
From the rest of the projects, we selected five projects for \unseenTest to allow
testing on projects with completely unknown structure and types.
We split the remaining 20 projects into train/validation/test sets in the
proportion 60-5-35, splitting along files (\ie, \emph{all} examples from one
source file are in the same set).
We call the test set obtained like this \seenTest.

\subsection{Quantitative Evaluation}
As a structured prediction
problem, there are multiple measures of performance on the task. In the
first part of \autoref{tbl:evaluation}, we report metrics when considering one 
placeholder at a time, \ie as if we are pasting a single variable identifier.
Accuracy reports the percent of correct predictions, MRR reports the mean reciprocal
rank of each prediction. We also measure type correctness, \ie the percent of
single-placeholder suggestions that yielded a suggestion of the correct type. In a
similar fashion, we present the results when pasting a full snippet. Now,
we perform structured prediction over all the placeholders within each snippet,
so we can now further compute exact match metrics over all the placeholders.
All the models reported here are using a log-bilinear model for computing
the context representation $\localContextRepr{\tok_k}$.
Using a GRU for computing $\localContextRepr{\tok_k}$ yielded slightly worse
results for all models.
We believe that this is due to optimization issues caused by the increased
depth of the network.

\begin{table}[t]
\caption{Evaluation of models. \unseenTest refers to projects were not part of the
train-test split. \seenTest refers to the test set containing projects
that have files in the training set.} \label{tbl:evaluation}
\resizebox{\textwidth}{!}{%
\scriptsize  
\begin{tabular}{@{}lrrrrrrrrrrr@{}} \toprule
& \multicolumn{5}{c}{\seenTest} && \multicolumn{5}{c}{\unseenTest} \\
 & \localmodel & \lblmodel & \grumodel & \grudfmodel & \hybridmodel && \localmodel & \lblmodel & \grumodel & \grudfmodel & \hybridmodel\\ \cmidrule{2-6} \cmidrule{8-12}
 \multicolumn{8}{l}{\underline{\textbf{Per Placeholder}}} \\
 Accuracy (\%) & 41.0 & 57.8 & 58.8& 56.2& \textbf{59.5}&& 27.9 & 55.2 & 52.8 & 51.3 & \textbf{56.2}\\
 MRR  & 0.562& 0.719& 0.723 & 0.701 & \textbf{0.727} && 0.476 & 0.709 & 0.683 & 0.664 & \textbf{0.710}\\
 Type Match (\%)  & 58.0& 68.4& 70.0& 67.8& \textbf{70.2}&& 47.5 & 64.6 & 62.8 & 61.7 & \textbf{65.4}\\
 \multicolumn{8}{l}{\underline{\textbf{Full Snippet Pasting}}} \\
 Accuracy (\%) & 47.4 & 57.9& 57.5& 55.3& \textbf{58.6}&& 31.7 & \textbf{54.9} & 49.9 & 49.3 & 54.5\\
 MRR  & 0.617 & 0.711& 0.712& 0.693& \textbf{0.716} && 0.491 & \textbf{0.709} & 0.666 & 0.655 & 0.700\\
 Ex Match (\%) & 20.5& 30.3& \textbf{31.3}& 28.8& 30.7 && 8.2 & 22.5 & 22.3 & 21.6 & \textbf{25.0}\\
 Type Match (\%) & 54.1& 67.3& 66.7& 64.5& \textbf{68.0}&& 45.4 & \textbf{62.7} & 56.8 & 56.4 & 61.2\\
 Type Ex Match (\%) & 32.0& 37.9& \textbf{39.3}& 36.1& 38.7&& 18.5 & \textbf{31.1} & 27.1 & 26.5 & 30.5\\
 \multicolumn{8}{l}{\underline{\textbf{Single Placeholder Same-Type Decisions}}} \\
 PR AUC & 0.543 & 0.819 & \textbf{0.835} & 0.806 & 0.830 && 0.494 & \textbf{0.849} & 0.839 & 0.833 & 0.833\\
 Precision@10\% & 87.0 & 96.5& \textbf{99.1} & 98.7 & 97.5 && 64.0 & \textbf{99.5} & 98.8 & 98.6 & 99.0\\
 \bottomrule
\end{tabular}
}\vspace{-1em}
\end{table}

Our results in \autoref{tbl:evaluation} show that \localmodel{} ---
as expected --- performs worse than all other models, indicating that our other
models learn valuable information from the provided usage contexts.
Somewhat surprisingly, our relatively simple \lblmodel{} already performs well.
On the other hand, \grudfmodel{} performs worse than models not taking the flow
of data into account.
We investigated this behavior more closely and found that the lexical context
models can often profit from observing the use of variables in other branches of
a conditional statement (\ie, peek at the \texttt{then} case when handling a
snippet in the \texttt{else} branch).
Consequently, \hybridmodel{}, which combines data flow information with all
usages always achieves high performance using both kinds of information.
Finally, to evaluate the need for type information, we run the experiments removing
all type information. This --- on average --- resulted to an 8\% reduced performance
on the \smartpaste task on all models.

\paragraph{Same-Type Decisions} 
So far, we considered the \smartpaste task where for each placeholder the neural networks
consider all variables in scope. However, if we assume that we know the desired type of
the placeholder, we can limit the set of suggestions.  The last set of metrics in \autoref{tbl:evaluation}
evaluate this scenario, \ie the suggestion performance within placeholders that have two or
more type-correct possible suggestions. In our dataset, there are on average 5.4 (median 2) same-type
variables in-scope per placeholder used in this evaluation. All networks (except \localmodel) achieve high precision-recall 
with a high AUC. This implies that our networks do \emph{not} just learn typing information.
Furthermore, for 10\% recall our best model achieves a precision of 99.1\%.
First, this suggests that these models can be used as a high-precision method for detecting bugs
caused by copy-pasting or porting that the code's existing type system would fail to catch.
Additionally, this indicates that our model have learned a probabilistic refinement of the
existing type system, \ie, that they can distinguish counters from other \texttt{int}
variables; file names from other \texttt{string}s; \etc

\paragraph{Generalization to new projects} 
Generalizing across a diverse set of source code projects with different domains
is an important challenge in machine learning. We repeat the evaluation using
the \unseenTest set stemming from projects that have no files in the training set.
The right side of \autoref{tbl:evaluation} shows that our models
still achieve good performance, although it is slightly lower compared to \seenTest,
especially when matching variable types. This is expected since the type lattice is
mostly unknown in \unseenTest.
We believe that some of the most important issues when transferring to new domains
is the fact that projects have significantly different type hierarchies and that
the vocabulary used (\eg by method names) is very different from the training projects.

\subsection{Qualitative Evaluation}
We show an example of the \smartpaste{} task in \autoref{fig:suggestions}, where
we can observe that the model learns to discriminate both among variables with
different types (\id{elements} of type \id{IHTMLElement} is not confused with
\id{string} variables) as well as assigning more fine-grained semantics
(\id{url} and \id{path} are treated separately) as implied by the results for
our same-type scenario above.

In \autoref{fig:nns}, we show placeholders that have highly similar usage context representations
\usageRepr{\tok}{\var}. Qualitatively, \autoref{fig:nns} and the visualizations in
\autoref{app:nns} suggest that the learned representations can be used as a learned
similarity metric for variable usage semantics. These representations learn protocols and
conventions such as ``after accessing \id{X}, we should access \id{Y}'' or the
need to conditionally check a property, as shown in \autoref{fig:nns}.

We observed a range of common problems.
Most notably, variables that are declared but not explicitly initialized (\eg as
a method parameter) cause the usage representation to be uninformative, grouping
all such declarations into the same representation.
The root cause is the limited information available in the context
representations. Local optima in ICM and UNK tokens also are common.

\begin{figure}[t]
\include{figures/fullsuggestion}
\caption{\label{fig:suggestions}\smartpaste suggestion on snippet of the \seenTest set.
\hybridmodel suggests all the red placeholders ($\lambda_6$ to $\lambda_{10}$) in the shaded area
($\lambda_5$ is a declaration). The probability for each placeholder is shown on the right.
Note that there are multiple \id{string} variables in scope (\id{url}, \id{pageUrl}, \id{path}).
However, \hybridmodel learns usage patterns (\eg \id{url} is a parameter of \id{IsFileUrl}) to assign
different representations to each variable usage. This allows us to discriminate between \id{path}, \id{url} and \id{pageUrl}.
The model ranks second the ground truth for $\lambda_9, \lambda_{10}$ suggesting \id{path} instead,
which nevertheless seems reasonable.
Variable names are \emph{not} used in the model, but are shown for convenience. Additional
visualizations are available in \autoref{app:fullsnippetviz}.
}
\end{figure}

\begin{figure}[t]
\input{figures/nearestneighbors}
\caption{\label{fig:nns}Placeholders (in black \blankplaceholder) with similar usage embeddings \usageRepr{\tok}{\var}.
Both \id{\_generatedCodeAnalysisFlagsOpt} and \id{symbolAndProjectId} implement the \id{Nullable}
interface. Note that the local context \lstinline{if(}\blankplaceholder\lstinline{.HasValue)}  is \emph{not}
used when computing \usageRepr{\tok}{\var} but data flow information of the other usages is used (marked in yellow).
In this example, the model learns a common representation for \id{Nullable}s that are assigned
and then conditionally used by accessing the \id{.HasValue} property.
The formatting of the snippets has been changed for space saving.
More examples can be found in \autoref{app:perplaceholderviz}.}
\end{figure}


\section{Related Work}
Our work builds upon the recent field of using machine learning for source code artifacts.
Recent research has lead to language models of code that try to model the whole
code. \citet{hindle2012naturalness,bhoopchand2016learning} model the code as a sequence of tokens,
while \citet{maddison2014structured,raychev2016probabilistic} model the syntax tree structure of code.
All the work on language models of code find that predicting variable and method identifiers is one of 
biggest challenges in the task. We are not aware of any models that attempt to use data flow information for variables.

Closest to our work is the work of \citet{allamanis2015suggesting}
who learn distributed representations of variables using all their usages
to predict their names.
However, they do not use data flow information and only consider semantically
equivalent renaming of variables ($\alpha$-renaming).
Finally, the work of \citet{raychev2015predicting} is also relevant, as it uses
a dependency network between variables.
However, all variable usages are deterministically known beforehand (as the code
is complete and remains unmodified), as in \citet{allamanis2014learning,allamanis2015suggesting}.

Our work is remotely related to work on program synthesis using sketches \citep{solar2008program}
and automated code transplantation \citep{barr2015automated}.
However, these approaches require a set of specifications (\eg input-output examples, test suites) to complete
the gaps, rather than statistics learned from big code. These approaches can be thought as complementary to ours,
since we learn to statistically complete the gaps
without any need for specifications, by learning common dataflow structure from code.

Our problem has also similarities with coreference resolution in NLP and methods for the
structured prediction of graphs and --- more commonly --- trees.  However, given the
different characteristics of the problems, such as the existence of exact execution path
information, we are not aware of any work
that would be directly relevant. Somewhat similar to our work, is the work of
\citet{clark2016improving}, who create a neural model that learns to rank pairs
of clusters of mentions to either merge them into a single co-reference entity or keep them apart.

\section{Discussion \& Conclusions}
Although source code is well understood and studied within other 
disciplines such as programming language research, it is a relatively
new domain for deep learning.
It presents novel opportunities compared to textual or perceptual data, as
its (local) semantics are well-defined and rich additional information can
be extracted using well-known, efficient program analyses.
On the other hand, integrating this wealth of structured information poses
an interesting challenge.
Our \smartpaste task exposes these opportunities, going beyond more simple
tasks such as code completion.
We consider it as a first proxy for the core challenge of learning the
\emph{meaning} of source code, as it requires to probabilistically refine
standard information included in type systems.

We see a wealth of opportunities in the research area.
To improve on our performance on the \smartpaste task, we want to extend
our models to additionally take identifier names into account, which are
obviously rich in information.
Similarly, we are interested in exploring more advanced tasks such as
bug finding, automatic code reviewing, \etc
\subsubsection*{Acknowledgments}
We would like to thank Alex Gaunt for his valuable comments
and suggestions.

\small
\bibliographystyle{abbrvnat}
\bibliography{bibliography}
\normalsize
\appendix
\section{Per Placeholder Suggestion Samples}
\label{app:perplaceholderviz}
Below we list a set of sample same-type decisions made when considering
one placeholder at a time. Some code comments and formatting have been
altered for typesetting reasons. The ground truth choice is underlined.

\begin{minipage}{\textwidth}
\textbf{Sample 1}
\begin{lstlisting}[frame=tlbr]
private static DataTable CreateDataTable(int cols, string colNamePrefix)
{
    var table = new DataTable();
    for (int i = 1; i <= cols; i++)
    {
        table.Columns.Add(new DataColumn() { ColumnName = colNamePrefix + (*\namedplaceholder{\#1}*),
                                             DefaultValue = (*\namedplaceholder{\#2}*) });
    }
    table.Rows.Add(table.NewRow());
    return table;
}
\end{lstlisting}
\namedplaceholder{\#1} \underline{\id{i}: 84\%}, \id{cols}: 16\%\\
\namedplaceholder{\#2} \underline{\id{i}: 53\%}, \id{cols}: 47\%\\
\end{minipage}

\begin{minipage}{\textwidth}
\textbf{Sample 2}
\begin{lstlisting}[frame=tlbr]
public void A_VectorClock_must_not_happen_before_itself()
{
    var clock1 = VectorClock.Create();
    var clock2 = VectorClock.Create();
    
    ((*\namedplaceholder{\#1}*) != (*\namedplaceholder{\#2}*)).Should().BeFalse();
}
\end{lstlisting}
\namedplaceholder{\#1} \underline{\id{clock1}: 44\%}, \id{clock2}: 56\%\\
\namedplaceholder{\#2} \id{clock1}: 9\%, \underline{\id{clock2}: 91\%}\\

\textbf{Sample 3}
\begin{lstlisting}[frame=tlbr]
public MergeHub(int perProducerBufferSize)
{
    if ((*\namedplaceholder{\#1}*) <= 0)
        throw new ArgumentException("Buffer size must be positive", nameof((*\namedplaceholder{\#2}*)));

    _perProducerBufferSize = perProducerBufferSize;
    DemandThreshold = perProducerBufferSize/2 + perProducerBufferSize%2;
    Shape = new SourceShape<T>(Out);
}
\end{lstlisting}
\namedplaceholder{\#1} \underline{\id{perProducerBufferSize}: 100\%}, \id{\_perProducerBufferSize}: 2e-4, \id{DemandThreshold}: 1e-6\\
\namedplaceholder{\#2} \underline{\id{perProducerBufferSize}: 100\%}, \id{\_perProducerBufferSize}: 3e-3, \id{DemandThreshold}: 2e-3\\
\end{minipage}

\begin{minipage}{\textwidth}
\textbf{Sample 4}
\begin{lstlisting}[frame=tlbr]
public Task UpdateRuntimeStatistics(SiloAddress siloAddress,
                                        SiloRuntimeStatistics siloStats)
{
    if (logger.IsVerbose)
        logger.Verbose("UpdateRuntimeStatistics from {0}", siloAddress);
    if (this.siloStatusOracle.GetApproximateSiloStatus(siloAddress) 
                                                    != SiloStatus.Active)
        return Task.CompletedTask;

    SiloRuntimeStatistics old;
    // Take only if newer.
    if (periodicStats.TryGetValue(siloAddress, out old) 
                        && old.DateTime > siloStats.DateTime)
        return Task.CompletedTask;

    (*\namedplaceholder{\#1}*)[siloAddress] = (*\namedplaceholder{\#2}*);
    NotifyAllStatisticsChangeEventsSubscribers(siloAddress, (*\namedplaceholder{\#3}*));
    return Task.CompletedTask;
}
\end{lstlisting}
\namedplaceholder{\#1} \underline{\id{periodicStats}: 94\%}, \id{PeriodicStats}: 6\%\\
\namedplaceholder{\#2} \underline{\id{siloStats}: 89\%}, \id{old}: 11\%\\
\namedplaceholder{\#3} \underline{\id{old}: 54\%}, \id{siloStats}: 46\%\\
\end{minipage}

\begin{minipage}{\textwidth}
\textbf{Sample 5}
\begin{lstlisting}[frame=tlbr]
public override BoundNode VisitLocal(BoundLocal node)
{
    LocalSymbol localSymbol = node.LocalSymbol;
    CheckAssigned(localSymbol, node.Syntax);

    if (localSymbol.IsFixed &&
        (this.(*\namedplaceholder{\#1}*).MethodKind == MethodKind.AnonymousFunction ||
            this.(*\namedplaceholder{\#2}*).MethodKind == MethodKind.LocalFunction) &&
        (*\namedplaceholder{\#3}*).Contains(localSymbol))
    {
        Diagnostics.Add(ErrorCode.ERR_FixedLocalInLambda,
                        new SourceLocation(node.Syntax), localSymbol);
    }
    return null;
}
\end{lstlisting}
\namedplaceholder{\#1} \underline{\id{currentMethodOrLambda}: 100\%}, \id{topLevelMethod}: 4e-3 \\
\namedplaceholder{\#2} \underline{\id{currentMethodOrLambda}: 100\%}, \id{topLevelMethod}: 2e-4 \\
\namedplaceholder{\#3} \id{\_writtenVariables}: 60\%,\underline{\id{\_capturedVariables}: 40\%} \\
\end{minipage}

\begin{minipage}{\textwidth}
\textbf{Sample 6}
\begin{lstlisting}[frame=tlbr]
private IDbContextServices InitializeServices()
{
    if ((*\namedplaceholder{\#1}*))
    {
        throw new InvalidOperationException(CoreStrings.RecursiveOnConfiguring);
    }
...
\end{lstlisting}
\namedplaceholder{\#1} \underline{\id{\_initializing}: 73\%}, \id{\_disposed}: 27\% \\
\end{minipage}

\begin{minipage}{\textwidth}
\textbf{Sample 7}
\begin{lstlisting}[frame=tlbr]
public static IMutableForeignKey GetOrAddForeignKey(
            [NotNull] this IMutableEntityType entityType,
            [NotNull] IReadOnlyList<IMutableProperty> properties,
            [NotNull] IMutableKey principalKey,
            [NotNull] IMutableEntityType principalEntityType)
{
    Check.NotNull((*\namedplaceholder{\#1}*), nameof((*\namedplaceholder{\#2}*)));

    return (*\namedplaceholder{\#3}*).FindForeignKey(properties, principalKey, (*\namedplaceholder{\#4}*))
            ?? (*\namedplaceholder{\#5}*).AddForeignKey(properties, principalKey, (*\namedplaceholder{\#6}*));
}
\end{lstlisting}
\namedplaceholder{\#1} \underline{\id{entityType}: 100\%}, \id{principalEntityType}: 4e-3 \\
\namedplaceholder{\#2} \underline{\id{entityType}: 100\%}, \id{principalEntityType}: 3e-4 \\
\namedplaceholder{\#3} \underline{\id{entityType}: 78\%}, \id{principalEntityType}: 22\% \\
\namedplaceholder{\#4} \underline{\id{principalEntityType}: 100\%}, \id{entityType}: 2e-3 \\
\namedplaceholder{\#5} \id{principalEntityType}: 60\%, \underline{\id{entityType}: 30\%} \\
\namedplaceholder{\#6} \id{entityType}: 99\%, \underline{\id{principalEntityType}: 1\%} \\
\end{minipage}

\begin{minipage}{\textwidth}
\textbf{Sample 8}
\begin{lstlisting}[frame=tlbr]
public string URL
{
    get
    {
        if ((*\namedplaceholder{\#1}*) == null)
        {
            // Read the URL into a string
            Stream stream = (Stream)m_dataObject.GetData(DataFormatsEx.URLFormat);
            StreamReader reader = new StreamReader(stream);

            using (reader)
            {
                (*\namedplaceholder{\#2}*) = reader.ReadToEnd().Trim((char)0);
            }
        }
        return (*\namedplaceholder{\#3}*);
    }
}
\end{lstlisting}
\namedplaceholder{\#1} \underline{\id{m\_url}: 90\%}, \id{m\_title}: 5\%, \id{URL}: 4\%, \id{Title}: 1\% \\
\namedplaceholder{\#2} \underline{\id{m\_url}: 84\%}, \id{Title}:13\%, \id{m\_title}: 1\%, \id{URL}: 1\%, \\
\namedplaceholder{\#3} \underline{\id{m\_url}: 99\%}, \id{m\_title}: 4e-3, \id{URL}: 3e-3, \id{Title}:6e-4 \\
\end{minipage}

\begin{minipage}{\textwidth}
\textbf{Sample 9}
\begin{lstlisting}[frame=tlbr]
internal static byte[] UrlEncodeToBytes(byte[] bytes, int offset, int count)
{
    if (bytes == null)
        throw new ArgumentNullException("bytes");

    int blen = bytes.Length;
    if ((*\namedplaceholder{\#1}*) == 0)
        return ArrayCache.Empty<byte>();

    if ((*\namedplaceholder{\#2}*) < 0 || (*\namedplaceholder{\#3}*) >= (*\namedplaceholder{\#4}*))
        throw new ArgumentOutOfRangeException("offset");
...      
\end{lstlisting}
\namedplaceholder{\#1} \underline{\id{blen}: 85\%}, \id{offset}: 9\%, \id{count}: 6\%\\
\namedplaceholder{\#2} \underline{\id{offset}: 43\%}, \id{blen}: 36\%, \id{count}: 21\%\\
\namedplaceholder{\#3} \underline{\id{offset}: 76\%}, \id{blen}: 13\%, \id{count}: 11\% \\
\namedplaceholder{\#4} \id{count}: 60\%, \id{offset}: 31\%,  \underline{\id{blen}: 10\%} \\
\end{minipage}

\begin{minipage}{\textwidth}
\textbf{Sample 10}
\begin{lstlisting}[frame=tlbr]
private static List<UsingDirectiveSyntax> AddUsingDirectives(
            CompilationUnitSyntax root, IList<UsingDirectiveSyntax> usingDirectives)
{
    // We need to try and not place the using inside of a directive if possible.
    var usings = new List<UsingDirectiveSyntax>();
    var endOfList = root.Usings.Count - 1;
    var startOfLastDirective = -1;
    var endOfLastDirective = -1;
    for (var i = 0; (*\namedplaceholder{\#1}*) < root.Usings.Count; (*\namedplaceholder{\#2}*)++)
    {
        if (root.Usings[(*\namedplaceholder{\#3}*)].GetLeadingTrivia()
                        .Any(trivia => trivia.IsKind(SyntaxKind.IfDirectiveTrivia)))
        {
            (*\namedplaceholder{\#4}*) = (*\namedplaceholder{\#5}*);
        }

        if (root.Usings[(*\namedplaceholder{\#6}*)].GetLeadingTrivia()
                        .Any(trivia => trivia.IsKind(SyntaxKind.EndIfDirectiveTrivia)))
        {
            (*\namedplaceholder{\#7}*) = (*\namedplaceholder{\#8}*);
        }
    }
...
\end{lstlisting}
\namedplaceholder{\#1} \underline{\id{i}: 98\%}, \id{endOfList}: 1\%, \id{startOfLastDirective}: 2e-3, \id{endOfLastDirective}: 3e-3  \\
\namedplaceholder{\#2} \underline{\id{i}: 99\%}, \id{endOfList}: 2e-3, \id{startOfLastDirective}: 6e-3, \id{endOfLastDirective}: 1e-3  \\
\namedplaceholder{\#3} \underline{\id{i}: 100\%}, \id{endOfList}: 1e-4, \id{startOfLastDirective}: 3e-4\%, \id{endOfLastDirective}: 5e-5  \\
\namedplaceholder{\#4} \id{endOfLastDirective}: 58\%, \underline{\id{startOfLastDirective}: 30\%}, \id{endOfList}: 1\%, \id{i}: 3e-3   \\
\namedplaceholder{\#5} \id{endOfLastDirective}: 77\%, \id{startOfLastDirective}: 12\%, \underline{\id{i}: 6\%}, \id{endOfList}: 5\%   \\
\namedplaceholder{\#6} \underline{\id{i}: 100\%}, \id{endOfList}: 1e-3, \id{startOfLastDirective}: 2e-4, \id{endOfLastDirective}: 5e-4  \\
\namedplaceholder{\#7} \underline{\id{endOfLastDirective}: 52\%}, \id{startOfLastDirective}: 37\%, \id{endOfList}: 1\%, \id{i}: 3e-3, \\
\namedplaceholder{\#8} \underline{\id{i}: 53\%}, \id{endOfLastDirective}: 27\%, \id{startOfLastDirective}: 13\%, \id{endOfList}: 7\% \\
\end{minipage}

\section{Nearest Neighbor of Usage Representations}
\label{app:nns}
Here we show pairs of nearest neighbors based on the cosine similarity of 
the learned representations \usageRepr{\tok}{\var}. Each placeholder $\tok$ is marked
as \blankplaceholder\xspace and all usages of $\var$ are marked in yellow (\ie \placeholder{variableName}).
Although names of variables are shown for convenience, they are \emph{not} used (only their
types --- if known --- is used).
 This is a set of hand-picked examples showing good
and bad examples. A brief description follows after each pair.
\newcommand{\explain}{$\triangleright$\xspace}

\begin{minipage}{\textwidth}
\textbf{Sample 1}
\begin{lstlisting}[frame=tlbr]
public void SetDateTime(string year, string month, string day)
{
    string (*\placeholder{time}*) = "";
    if (year.Contains(":"))
    {
        (*\blankplaceholder*) = year;
        year = DateTime.Now.Year.ToString();
        TimeInfo = true;
    }

    DateTime = DateTime.Parse(string.Format("{0}/{1}/{2} {3}", year,
                                 month, day, (*\placeholder{time}*)));
    DateTime = DateTime.ToLocalTime();
}
\end{lstlisting}
\begin{lstlisting}[frame=tlbr]
public void MakeMultiDirectory(string dirName)
{
    string (*\placeholder{path}*) = "";
    string[] dirs = dirName.Split('/');
    foreach (string dir in dirs)
    {
        if (!string.IsNullOrEmpty(dir))
        {
            (*\blankplaceholder*) = URLHelpers.CombineURL(path, dir);
            MakeDirectory(URLHelpers.CombineURL(Options.Account.FTPAddress, (*\placeholder{path}*)));
        }
    }

    WriteOutput("MakeMultiDirectory: " + dirName);
}
\end{lstlisting}
\explain Usage context where a string has been initialized to blank but may
be reassigned before it is used.
\end{minipage}

\begin{minipage}{\textwidth}
\textbf{Sample 2}
\begin{lstlisting}[frame=tlbr]
...
FtpWebRequest (*\placeholder{request}*) = (FtpWebRequest)WebRequest.Create(url);
(*\blankplaceholder*).Proxy = Options.ProxySettings;
(*\placeholder{request}*).Method = WebRequestMethods.Ftp.ListDirectory;
(*\placeholder{request}*).Credentials = new NetworkCredential(Options.Account.Username,
             Options.Account.Password);
(*\placeholder{request}*).KeepAlive = false;
(*\placeholder{request}*).Timeout = 10000;
(*\placeholder{request}*).UsePassive = !Options.Account.IsActive;

using (WebResponse response = (*\placeholder{request}*).GetResponse()) {
...
\end{lstlisting}
\begin{lstlisting}[frame=tlbr]
...
FtpWebRequest (*\placeholder{request}*) = (FtpWebRequest)WebRequest.Create(url);
(*\blankplaceholder*).Proxy = Options.ProxySettings;
(*\placeholder{request}*).Method = WebRequestMethods.Ftp.RemoveDirectory;
(*\placeholder{request}*).Credentials = new NetworkCredential(Options.Account.Username,
             Options.Account.Password);
(*\placeholder{request}*).KeepAlive = false;

(*\placeholder{request}*).GetResponse();
...
\end{lstlisting}
\explain Similar protocols when using an object.
\end{minipage}

\begin{minipage}{\textwidth}
\textbf{Sample 3}
\begin{lstlisting}[frame=tlbr]
...
var (*\placeholder{addMethod}*) = @event.AddMethod;
Assert.Equal(voidType, (*\blankplaceholder*).ReturnType);
Assert.True((*\placeholder{addMethod}*).ReturnsVoid);
Assert.Equal(1, (*\placeholder{addMethod}*).ParameterCount);
Assert.Equal(eventType, (*\placeholder{addMethod}*).ParameterTypes.Single());
...
\end{lstlisting}
\begin{lstlisting}[frame=tlbr]
...
var (*\placeholder{removeMethod}*) = @event.RemoveMethod;
Assert.Equal(voidType, (*\blankplaceholder*).ReturnType);
Assert.True((*\placeholder{removeMethod}*).ReturnsVoid);
Assert.Equal(1, (*\placeholder{removeMethod}*).ParameterCount);
Assert.Equal(eventType, (*\placeholder{removeMethod}*).ParameterTypes.Single());
...
\end{lstlisting}
\explain These two placeholders have --- by definition --- identical representations.
\end{minipage}

\begin{minipage}{\textwidth}
\textbf{Sample 4}
\begin{lstlisting}[frame=tlbr]
...
int (*\placeholder{index}*) = flpHotkeys.Controls.GetChildIndex(Selected);

int newIndex;

if ((*\blankplaceholder*) == 0)
{
    newIndex = flpHotkeys.Controls.Count - 1;
}
else
{
    newIndex = (*\placeholder{index}*) - 1;
}

flpHotkeys.Controls.SetChildIndex(Selected, newIndex);
manager.Hotkeys.Move((*\placeholder{index}*), newIndex);
...
\end{lstlisting}
\begin{lstlisting}[frame=tlbr]
...
if (Selected != null && flpHotkeys.Controls.Count > 1)
{
    int (*\placeholder{index}*) = flpHotkeys.Controls.GetChildIndex(Selected);

    int newIndex;

    if ((*\blankplaceholder*) == flpHotkeys.Controls.Count - 1)
    {
        newIndex = 0;
    }
    else
    {
        newIndex = (*\placeholder{index}*) + 1;
    }

    flpHotkeys.Controls.SetChildIndex(Selected, newIndex);
    manager.Hotkeys.Move((*\placeholder{index}*), newIndex);
...
\end{lstlisting}

\end{minipage}

\begin{minipage}{\textwidth}
\textbf{Sample 5}
\begin{lstlisting}[frame=tlbr]
int index = flpHotkeys.Controls.GetChildIndex(Selected);
int (*\placeholder{newIndex}*);
if (index == 0)
{
    (*\blankplaceholder*) = flpHotkeys.Controls.Count - 1;
}
else
{
    (*\placeholder{newIndex}*) = index - 1;
}
flpHotkeys.Controls.SetChildIndex(Selected, (*\placeholder{newIndex}*));
manager.Hotkeys.Move(index, (*\placeholder{newIndex}*));
\end{lstlisting}
\begin{lstlisting}[frame=tlbr]
int index = flpHotkeys.Controls.GetChildIndex(Selected);
int (*\placeholder{newIndex}*);
if (index == 0)
{
    (*\placeholder{newIndex}*) = flpHotkeys.Controls.Count - 1;
}
else
{
    (*\blankplaceholder*) = index - 1;
}
flpHotkeys.Controls.SetChildIndex(Selected, (*\placeholder{newIndex}*));
manager.Hotkeys.Move(index, (*\placeholder{newIndex}*));
\end{lstlisting}
\explain Because of the dataflow, these two placeholders (one in each branch of the \id{if-else})
have identical representations in the dataflow model, and have very similar representations
in other models.
\end{minipage}

\begin{minipage}{\textwidth}
\textbf{Sample 6}
\begin{lstlisting}[frame=tlbr]
(*\placeholder{\_generatedCodeAnalysisFlagsOpt}*) = generatedCodeAnalysisFlagsOpt;
...
context.RegisterCompilationStartAction(this.OnCompilationStart);

if ((*\blankplaceholder*).HasValue)
{
    // Configure analysis on generated code.
    context.ConfigureGeneratedCodeAnalysis((*\placeholder{\_generatedCodeAnalysisFlagsOpt}*).Value);
}
...
\end{lstlisting}
\begin{lstlisting}[frame=tlbr]
...
var (*\placeholder{symbolAndProjectId}*) = await definition.TryRehydrateAsync(
    _solution, _cancellationToken).ConfigureAwait(false);

if (!(*\blankplaceholder*).HasValue)
{
    return;
}

lock (_gate)
{
    _definitionMap[definition] = (*\placeholder{symbolAndProjectId}*).Value;
}
...
\end{lstlisting}
\explain Our model learns a similar representation for the placeholder between the
locations where a
\id{Nullable} variable is assigned and used, which corresponds to a check on the
\id{.HasValue} property.
\end{minipage}

\begin{minipage}{\textwidth}
\textbf{Sample 7}
\begin{lstlisting}[frame=tlbr]
var analyzers = new DiagnosticAnalyzer[] { new ConcurrentAnalyzer(typeNames) };
var expected = new DiagnosticDescription[typeCount];
for (int (*\placeholder{i}*) = 0; (*\blankplaceholder*) < typeCount; (*\placeholder{i}*)++)
{
    var typeName = $"C{(*\placeholder{i}*) + 1}";
    expected[(*\placeholder{i}*)] = Diagnostic(ConcurrentAnalyzer.Descriptor.Id, typeName)
        .WithArguments(typeName)
        .WithLocation((*\placeholder{i}*) + 2, 7);
}
\end{lstlisting}
\begin{lstlisting}[frame=tlbr]
var builder = new StringBuilder();
var typeCount = 100;
var typeNames = new string[typeCount];
for (int (*\placeholder{i}*) = 1; (*\blankplaceholder*) <= typeCount; (*\placeholder{i}*)++)
{
    var typeName = $"C{(*\placeholder{i}*)}";
    typeNames[(*\placeholder{i}*) - 1] = typeName;
    builder.Append($"\r\nclass {typeName} {{ }}");
}
\end{lstlisting}
\explain The model learns --- unsurprisingly --- a very similar representation of the loop control variable \id{i}
at the location of the bound check. Generalizing over varying loops.
\end{minipage}

\begin{minipage}{\textwidth}
\textbf{Sample 8}
\begin{lstlisting}[frame=tlbr]
...
if (!(*\blankplaceholder*))
{
    if (disposeManagedResources)
    {
        _resizerControl.SizerModeChanged += 
                    new SizerModeEventHandler(resizerControl_SizerModeChanged);
        _resizerControl.Resized -= new EventHandler(resizerControl_Resized);
        _dragDropController.Dispose();
    }

     (*\placeholder{\_disposed}*) = true;
}
...
\end{lstlisting}
\begin{lstlisting}[frame=tlbr]
...
if (!(*\blankplaceholder*))
{
    _enableRealTimeWordCount = Settings.GetBoolean(SHOWWORDCOUNT, false);
    (*\placeholder{\_enableRealTimeWordCountInit}*) = true;
}
return _enableRealTimeWordCount;
...
\end{lstlisting}
\explain Similar representations for booleans that will be assigned to \id{true} within a branch.
\end{minipage}

\begin{minipage}{\textwidth}
\textbf{Sample 9}
\begin{lstlisting}[frame=tlbr]
...
SmartContentSelection (*\placeholder{selection}*) = EditorContext.Selection as SmartContentSelection;
if ((*\blankplaceholder*) != null)
{
    return (*\placeholder{selection}*).HTMLElement.sourceIndex == HTMLElement.sourceIndex;
}
else
{
    return false;
}
...
\end{lstlisting}
\begin{lstlisting}[frame=tlbr]
...
foreach (LiveClipboardFormat format in formats)
{
 ContentSourceInfo (*\placeholder{contentSource}*) = FindContentSourceForLiveClipboardFormat(format);
 if ((*\blankplaceholder*) != null)
    return (*\placeholder{contentSource}*);
}
...
\end{lstlisting}
\explain Representation for elements that will be returned but only one one path.
\end{minipage}

\begin{minipage}{\textwidth}
\textbf{Sample 10}
\begin{lstlisting}[frame=tlbr]
...
Rectangle (*\placeholder{elementRect}*) = ElementRectangle;
_resizerControl.VirtualLocation = new Point(
            (*\placeholder{elementRect}*).X - ResizerControl.SIZERS_PADDING,
            (*\blankplaceholder*).Y - ResizerControl.SIZERS_PADDING);  
...
\end{lstlisting}
\begin{lstlisting}[frame=tlbr]
...
Rectangle (*\placeholder{rect}*) = CalculateElementRectangleRelativeToBody(HTMLElement);
IHTMLElement body = (HTMLElement.document as IHTMLDocument2).body;

_dragBufferControl.VirtualSize = new Size(body.offsetWidth, body.offsetHeight);
_dragBufferControl.VirtualLocation = new Point(-(*\placeholder{rect}*).X, -(*\blankplaceholder*).Y);
_dragBufferControl.Visible = true;
...
\end{lstlisting}
\explain Protocol of accesses for \id{Rectangle} objects. After \id{X} has been accessed then the variable has the same
representation (implied that \id{Y} is highly likely to be accessed next)
\end{minipage}

\begin{minipage}{\textwidth}
\textbf{Sample 11}
\begin{lstlisting}[frame=tlbr]
...
if ((*\placeholder{parameters}*) != null)
{
    for (int i = 0; i < (*\placeholder{parameters}*).Length; i += 2)
    {
        string name = (*\blankplaceholder*)[i];
        string val = (*\placeholder{parameters}*)[i + 1];
        if (!cullMissingValues || (val != null && val != string.Empty))
            Add(name, val);
    }
}
...
\end{lstlisting}
\begin{lstlisting}[frame=tlbr]
...
string[] (*\placeholder{refParams}*) = value.Split(commaSeparator);

if ((*\placeholder{refParams}*).Length != 2 || string.IsNullOrEmpty((*\placeholder{refParams}*)[0]) 
                    || string.IsNullOrEmpty((*\placeholder{refParams}*)[1]))
    throw new ArgumentException("Reference path is invalid.");

ModuleName = (*\blankplaceholder*)[0];
ResourceId = int.Parse((*\placeholder{refParams}*)[1]);

referencePath = value;
...
\end{lstlisting}
\explain Similar representation for first array access after bound checks.
\end{minipage}

\begin{minipage}{\textwidth}
\textbf{Sample 12}
\begin{lstlisting}[frame=tlbr]
...
public static string GetHostName(string (*\placeholder{url}*))
{
    if (!IsUrl((*\blankplaceholder*)))
        return null;
    return new Uri((*\placeholder{url}*)).Host;
}
...
\end{lstlisting}
\begin{lstlisting}[frame=tlbr]
...
public static bool IsUrlLinkable(string (*\placeholder{url}*))
{
    if (UrlHelper.IsUrl((*\blankplaceholder*)))
    {
        Uri uri = new Uri((*\placeholder{url}*));
        foreach (string scheme in NonlinkableSchemes)
            if (uri.Scheme == scheme)
                return false;

    }
    return true;
}
...
\end{lstlisting}
\explain Similar representations because of learned pattern when parsing URIs.
\end{minipage}

\begin{minipage}{\textwidth}
\textbf{Sample 13}
\begin{lstlisting}[frame=tlbr]
...
private void WriteEntry(string message, string (*\placeholder{category}*), string stackTrace)
{
    //	Obtain the DateTime the message reached us.
    DateTime dateTime = DateTime.Now;

    //	Default the message, as needed.
    if (message == null || message.Length == 0)
        message = "[No Message]";

    //	Default the category, as needed.
    if ((*\placeholder{category}*) == null || (*\placeholder{category}*).Length == 0)
        (*\blankplaceholder*) = "None";

    int seqNum = Interlocked.Increment(ref sequenceNumber);

    DebugLogEntry logEntry = new DebugLogEntry(facility, processId, seqNum,
                                               dateTime, message, (*\placeholder{category}*), stackTrace);
...
\end{lstlisting}
\begin{lstlisting}[frame=tlbr]
...
private void WriteEntry(string (*\placeholder{message}*), string category, string stackTrace)
{
    //	Obtain the DateTime the message reached us.
    DateTime dateTime = DateTime.Now;

    //	Default the message, as needed.
    if ((*\placeholder{message}*) == null || (*\placeholder{message}*).Length == 0)
        (*\blankplaceholder*) = "[No Message]";

    //	Default the category, as needed.
    if (category == null || category.Length == 0)
        category = "None";

    int seqNum = Interlocked.Increment(ref sequenceNumber);

    DebugLogEntry logEntry = new DebugLogEntry(facility, processId, seqNum,
                                               dateTime, (*\placeholder{message}*), category, stackTrace);
...
\end{lstlisting}
\explain The variables \id{message} and \id{category} (in the same snippet of code) have
similar representations. This is a source of confusion for our models.
\end{minipage}

\begin{minipage}{\textwidth}
\textbf{Sample 14}
\begin{lstlisting}[frame=tlbr]
...
foreach (string (*\placeholder{file}*) in files)
{
    string[] chunks = (*\blankplaceholder*).Split(INTERNAL_EXTERNAL_SEPARATOR);
    switch (chunks[0])
    {
...
\end{lstlisting}
\begin{lstlisting}[frame=tlbr]
...
Uri uri = new Uri(url);
foreach (string (*\placeholder{scheme}*) in NonlinkableSchemes)
    if (uri.Scheme == (*\blankplaceholder*))
        return false;
...
\end{lstlisting}
\explain Limited context (\eg only declaration) causes variables to have similar
usage representations. In the examples \id{file} and \id{scheme} are defined and used
only once. This is a common source of confusion for our models.
\end{minipage}

\section{Full Snippet Pasting Samples}
\label{app:fullsnippetviz}
Below we present some of the suggestions when using
the full \smartpaste structured prediction. The variables shown
at each placeholder correspond to the ground truth. Underlined
tokens represent UNK tokens. The top three allocations are shown
as well as the ground truth (if it is \emph{not} in the top 3 suggestions).
Red placeholders are the placeholders that need to be filled in when pasting.
All other placeholders are marked in superscript next to the relevant
variable.

\begin{minipage}{\textwidth}
\textbf{Sample 1}
\begin{lstlisting}[xleftmargin=0cm,frame=tlbr]
...
(*\hlplacehld{charsLeft}{1}*) = 0;
while ((*\hlplacehld{p}{2}*).IsRightOf((*\hlplacehld{selection}{3}*).Start))
{
    (*\hlplacehld{charsLeft}{4}*)++;
    (*\hlplacehld{p}{5}*).MoveUnit(_MOVEUNIT_ACTION.MOVEUNIT_PREVCHAR);
}
...
\end{lstlisting}
\begin{description}
\item[\hlreflarge{1}] \underline{\id{charsLeft}: 87\%}, \id{movesRight}: 8\%, \id{p}: 5\%
\item[\hlreflarge{2}] \underline{\id{p}: 96\%}, \id{selection}: 4\%, \id{bounds}: 1e-3
\item[\hlreflarge{3}] \underline{\id{selection}: 89\%}, \id{bounds}: 1\%, \id{p}: 8e-3
\item[\hlreflarge{4}] \id{movesRight}: 66\%, \underline{\id{charsLeft}: 16\%}, \id{p}: 1\%
\item[\hlreflarge{5}] \underline{\id{p}: 83\%}, \id{selection}: 11\%, \id{bounds}: 6\%
\end{description}
\end{minipage}

\begin{minipage}{\textwidth}
\textbf{Sample 2}
\begin{lstlisting}[xleftmargin=0cm,frame=tlbr]
...
HttpWebResponse response(*\hlref{0}*) = null;
XmlDocument xmlDocument(*\hlref{1}*) = new XmlDocument();
try
{
    using (Blog blog(*\hlref{3}*) = new Blog((*\hlplacehld{\_blogId}{4}*)))
        (*\hlplacehld{response}{5}*) = (*\hlplacehld{blog}{6}*).SendAuthenticatedHttpRequest((*\hlplacehld{notificationUrl}{7}*), 10000);

    // parse the results
    (*\hlplacehld{xmlDocument}{8}*).Load((*\hlplacehld{response}{9}*).GetResponseStream());
}
catch (Exception)
{
    throw;
}
finally
{
    if ((*\hlplacehld{response}{10}*) != null)
        (*\hlplacehld{response}{11}*).Close();
}
...
\end{lstlisting}
\begin{description}
\item[\hlreflarge{4}] \id{\_hostBlogId}: 12\%, \id{BlogId}: 10\%, \id{\_buttonId}: 10\%, \underline{\id{\_blogId}: 1\%}
\item[\hlreflarge{5}] \underline{\id{response}: 86\%}, \id{xmlDocument}: 5\%, \id{notificationUrl}: 3\%
\item[\hlreflarge{6}] \id{xmlDocument}: 84\%, \underline{\id{blog}: 12\%}, \id{response}: 2\%
\item[\hlreflarge{7}] \id{NotificationPollingTime}: 95\%, \id{CONTENT\_DISPLAY\_SIZE}: 2\%, \underline{\id{notificationUrl}: 1\%}
\item[\hlreflarge{8}] \underline{\id{xmlDocument}: 100\%}, \id{response}: 9e-4, \id{\_buttonDescription}: 4e-4 
\item[\hlreflarge{9}] \underline{\id{response}: 65\%}, \id{xmlDocument}: 30\%, \id{\_hostBlogId}: 4\%
\item[\hlreflarge{10}] \underline{\id{response}: 90\%}, \id{\_blogId}: 3\%, \id{CurrentImage}: 9e-3
\item[\hlreflarge{11}] \underline{\id{response}: 98\%}, \id{\_settingKey}: 1\%, \id{xmlDocument}: 9e-3
\end{description}
\end{minipage}

\begin{minipage}{\textwidth}
\textbf{Sample 3}
\begin{lstlisting}[xleftmargin=0cm,frame=tlbr]
...
protected override void Dispose(bool disposing(*\hlref{1}*))
{
    if ((*\hlplacehld{disposing}{2}*))
    {
        if ((*\hlplacehld{components}{3}*) != null)
            (*\hlplacehld{components}{4}*).Dispose();
    }
    base.Dispose((*\hlplacehld{disposing}{5}*));
}
...
\end{lstlisting}
\begin{description}
\item[\hlreflarge{2}] \underline{\id{disposing}: 100\%}, \id{commandIdentifier}: 4e-4, \id{components}: 1e-4
\item[\hlreflarge{3}] \underline{\id{components}: 100\%}, \id{disposing}: 3e-5, \id{commandIdentifier}: 2e-5
\item[\hlreflarge{4}] \underline{\id{components}: 100\%}, \id{disposing}: 9e-7, \id{CommandIdentifier}: 6e-9
\item[\hlreflarge{5}] \underline{\id{disposing}: 100\%}, \id{components}: 3e-5, \id{CommandIdentifier}: 2e-5
\end{description}
\end{minipage}

\begin{minipage}{\textwidth}
\textbf{Sample 4}
\begin{lstlisting}[xleftmargin=0cm,frame=tlbr]
...
tmpRange(*\hlref{1}*).Start.MoveAdjacentToElement(startStopParent(*\hlref{2}*),
                                         _ELEMENT_ADJACENCY.ELEM_ADJ_BeforeBegin);
if (tmpRange(*\hlref{3}*).IsEmptyOfContent())
{
    tmpRange(*\hlref{4}*).Start.MoveToPointer(selection(*\hlref{5}*).End);
    IHTMLElement endStopParent(*\hlref{6}*) = tmpRange(*\hlref{7}*).Start.GetParentElement(stopFilter(*\hlref{8}*));

    if ((*\hlplacehld{endStopParent}{9}*) != null 
            && (*\hlplacehld{startStopParent}{10}*).sourceIndex == (*\hlplacehld{endStopParent}{11}*).sourceIndex)
    {
        (*\hlplacehld{tmpRange}{12}*).Start
                .MoveAdjacentToElement((*\hlplacehld{endStopParent}{13}*),
                                             _ELEMENT_ADJACENCY.ELEM_ADJ_BeforeEnd);
        if ((*\hlplacehld{tmpRange}{14}*).IsEmptyOfContent())
        {
            (*\hlplacehld{tmpRange}{15}*).MoveToElement((*\hlplacehld{endStopParent}{16}*), true);
            if ((*\hlplacehld{maximumBounds}{17}*).InRange((*\hlplacehld{tmpRange}{18}*)) 
                                        && !((*\hlplacehld{endStopParent}{19}*) is IHTMLTableCell))
            {
                (*\hlplacehld{deleteParentBlock}{20}*) = true;
            }
        }
    }
}
...
\end{lstlisting}
\begin{description}
\item[\hlreflarge{9}] \id{startStopParent}: 97\%, \id{styleTagId}: 1\%, \id{tmpRange}: 1\%, \underline{\id{endStopParent}: 3e-3}
\item[\hlreflarge{10}] \underline{\id{startStopParent}: 100\%}, \id{tmpRange}: 2e-4, \id{maximumBounds}: 3e-5
\item[\hlreflarge{11}] \id{startStopParent}: 100\%, \id{styleTagId}: 2e-3, \underline{\id{endStopParent}: 1e-3}
\item[\hlreflarge{12}] \underline{\id{tmpRange}: 99\%}, \id{selection}: 9e-3, \id{startStopParent}: 2e-3
\item[\hlreflarge{13}] \id{startStopParent}: 96\%, \id{tmpRange}: 2\%, \underline{\id{endStopParent}: 1\%}
\item[\hlreflarge{14}] \underline{\id{tmpRange}: 98\%}, \id{selection}: 1\%, \id{maximumBounds}: 1\%
\item[\hlreflarge{15}] \underline{\id{tmpRange}: 98\%}, \id{selection}: 2\%, \id{maximumBounds}: 4e-3
\item[\hlreflarge{16}] \id{startStopParent}: 43\%, \id{styleTagId}: 29\%, \underline{\id{endStopParent}: 21\%}
\item[\hlreflarge{17}] \id{tmpRange}: 70\%, \id{selection}: 14\%, \underline{\id{maximumBounds}: 8\%}
\item[\hlreflarge{18}] \id{styleTagId}: 84\%, \underline{\id{tmpRange}: 5\%}, \id{selection}: 5\%
\item[\hlreflarge{19}] \id{startStopParent}: 98\%, \underline{\id{endStopParent}: 1\%}, \id{styleTagId}: 9e-3
\item[\hlreflarge{20}] \underline{\id{deleteParentBlock}: 90\%}, \id{startStopParent}: 4\%, \id{selection}: 3\%
\end{description}
\end{minipage}

\begin{minipage}{\textwidth}
\textbf{Sample 5}
\begin{lstlisting}[xleftmargin=0cm,frame=tlbr]
...
public static void GetImageFormat(string srcFileName(*\hlref{1}*), out string extension(*\hlref{2}*),
                                  out ImageFormat imageFormat(*\hlref{3}*))
{
    (*\hlplacehld{extension}{4}*) = Path.GetExtension((*\hlplacehld{srcFileName}{5}*))
                                            .ToLower(CultureInfo.InvariantCulture);
    if ((*\hlplacehld{extension}{6}*) == ".jpg" || (*\hlplacehld{extension}{7}*) == ".jpeg")
    {
        (*\hlplacehld{imageFormat}{8}*) = ImageFormat.Jpeg;
        (*\hlplacehld{extension}{9}*) = ".jpg";
    }
    else if ((*\hlplacehld{extension}{10}*) == ".gif")
    {
        (*\hlplacehld{imageFormat}{11}*) = ImageFormat.Gif;
    }
    else
    {
        (*\hlplacehld{imageFormat}{12}*) = ImageFormat.Png;
        (*\hlplacehld{extension}{13}*) = ".png";
    }
}
...
\end{lstlisting}
\begin{description}
\item[\hlreflarge{4}] \underline{\id{extension}: 64\%}, \id{imageFormat}: 36\%, \id{JPEG\_QUALITY}: 1e-4
\item[\hlreflarge{5}] \id{extension}: 98\%, \underline{\id{srcFileName}: 1\%}, \id{imageFormat}: 1e-3
\item[\hlreflarge{6}] \underline{\id{extension}: 97\%}, \id{imageFormat}: 1\%, \id{srcFileName}: 3e-4
\item[\hlreflarge{7}] \underline{\id{extension}: 75\%}, \id{JPG}: 4\%, \id{GIF}: 4\%
\item[\hlreflarge{8}] \underline{\id{imageFormat}: 100\%}, \id{extension}: 1e-5, \id{JPEG\_QUALITY}: 2e-6
\item[\hlreflarge{9}] \underline{\id{extension}: 93\%}, \id{imageFormat}: 2\%, \id{JPEG}: 9e-3
\item[\hlreflarge{10}] \underline{\id{extension}: 52\%}, \id{imageFormat}: 15\%, \id{ICO}: 6\%, \id{JPG}: 6\%, \id{GIF}: 6\%
\item[\hlreflarge{11}] \underline{\id{imageFormat}: 100\%}, \id{extension}: 4e-4, \id{JPEG\_QUALITY}: 1e-5
\item[\hlreflarge{12}] \underline{\id{imageFormat}: 99\%}, \id{JPEG\_QUALITY}: 4e-3, \id{extension}: 2e-3
\item[\hlreflarge{13}] \underline{\id{extension}: 66\%}, \id{JPG}: 6\%, \id{ICO}: 6\%, \id{GIF}: 6\%, \id{BMP}: 6\%
\end{description}
\end{minipage}

\begin{minipage}{\textwidth}
\textbf{Sample 6}
\begin{lstlisting}[xleftmargin=0cm,frame=tlbr]
...
BitmapData destBitmapData(*\hlref{1}*) = scaledBitmap(*\hlref{2}*).LockBits(
        new Rectangle(0, 0, destWidth(*\hlref{3}*), destHeight(*\hlref{4}*)),
        ImageLockMode.WriteOnly, scaledBitmap(*\hlref{5}*).PixelFormat);
try
{
    byte* s0(*\hlref{6}*) = (byte*)(*\hlplacehld{sourceBitmapData}{7}*).Scan0.ToPointer();
    int sourceStride(*\hlref{8}*) = (*\hlplacehld{sourceBitmapData}{9}*).Stride;
    byte* d0(*\hlref{10}*) = (byte*)(*\hlplacehld{destBitmapData}{11}*).Scan0.ToPointer();
    int destStride(*\hlref{12}*) = (*\hlplacehld{destBitmapData}{13}*).Stride;

    for (int y(*\hlref{14}*) = 0; y(*\hlref{15}*) < destHeight(*\hlref{16}*); y(*\hlref{17}*)++)
    {
        byte* d(*\hlref{18}*) = d0(*\hlref{19}*) + y(*\hlref{20}*) * destStride(*\hlref{21}*);
        byte* sRow(*\hlref{22}*) = s0(*\hlref{23}*) + ((int)(y(*\hlref{24}*) * yRatio(*\hlref{25}*)) 
                                + yOffset(*\hlref{26}*)) * sourceStride(*\hlref{27}*) + xOffset(*\hlref{28}*);
...
\end{lstlisting}
\begin{description}
\item[\hlreflarge{7}] \underline{\id{sourceBitmapData}: 72\%}, \id{destBitmapData}: 28\%, \id{bitmap}: 2e-6
\item[\hlreflarge{9}] \underline{\id{sourceBitmapData}: 90\%}, \id{destBitmapData}: 10\%, \id{bitmap}: 1e-4
\item[\hlreflarge{11}] \id{sourceBitmapData}: 75\%, \underline{\id{destBitmapData}: 25\%}, \id{s0}: 1e-5
\item[\hlreflarge{13}] \id{sourceBitmapData}: 83\%, \underline{\id{destBitmapData}: 17\%}, \id{bitmap}: 3e-4
\end{description}
\end{minipage}

\begin{minipage}{\textwidth}
\textbf{Sample 7}
\begin{lstlisting}[xleftmargin=0cm,frame=tlbr]
...
private static Stream GetStreamForUrl(string url(*\hlref{1}*), string pageUrl(*\hlref{2}*),
                                                             IHTMLElement element(*\hlref{3}*))
{
    if (UrlHelper.IsFileUrl(url(*\hlref{4}*)))
    {
        string path(*\hlref{5}*) = new Uri((*\hlplacehld{url}{6}*)).LocalPath;
        if (File.Exists((*\hlplacehld{path}{7}*)))
        {
            return File.OpenRead((*\hlplacehld{path}{8}*));
        }
        else
        {
            if (ApplicationDiagnostics.AutomationMode)
                Trace.WriteLine("File " + (*\hlplacehld{url}{9}*) + " not found");
            else
                Trace.Fail("File " + (*\hlplacehld{url}{10}*) + " not found");
            return null;
        }
    }
    else if (UrlHelper.IsUrlDownloadable(url(*\hlref{11}*)))
    {
        return HttpRequestHelper.SafeDownloadFile(url(*\hlref{12}*));
    }
    else
    {
...
\end{lstlisting}
\begin{description}
\item[\hlreflarge{6}] \underline{\id{url}: 96\%}, \id{element}: 2\%, \id{pageUrl}: 1\%
\item[\hlreflarge{7}] \underline{\id{path}: 86\%}, \id{url}: 14\%, \id{element}: 1e-3
\item[\hlreflarge{8}] \underline{\id{path}: 99\%}, \id{url}: 1\%, \id{pageUrl}: 4e-5
\item[\hlreflarge{9}] \id{path}: 97\%, \underline{\id{url}: 2\%}, \id{pagrUrl}: 4e-3\%
\item[\hlreflarge{10}] \id{path}: 67\%, \underline{\id{url}: 24\%}, \id{pageUrl}: 5\%
\end{description}
\end{minipage}

\begin{minipage}{\textwidth}
\textbf{Sample 8}
\begin{lstlisting}[xleftmargin=0cm,frame=tlbr]
...
public static void ApplyAlphaShift(Bitmap bitmap(*\hlref{1}*), double alphaPercentage(*\hlref{2}*))
{
    for (int y(*\hlref{3}*) = 0; y(*\hlref{4}*) < bitmap(*\hlref{5}*).Height; y(*\hlref{6}*)++)
    {
        for (int x(*\hlref{7}*) = 0; x(*\hlref{8}*) < bitmap(*\hlref{9}*).Width; x(*\hlref{10}*)++)
        {
            Color c(*\hlref{11}*) = bitmap.GetPixel(x(*\hlref{12}*), y(*\hlref{13}*));
            if ((*\hlplacehld{c}{14}*).(*\underline{\id{A}}*) > 0) //never make transparent pixels non-transparent
            {
                int newAlphaValue(*\hlref{15}*) = (int)((*\hlplacehld{c}{16}*).(*\underline{\id{A}}*) * (*\hlplacehld{alphaPercentage}{17}*));
                //value must be between 0 and 255
                (*\hlplacehld{newAlphaValue}{18}*) = Math.Max(0, Math.Min(255, (*\hlplacehld{newAlphaValue}{19}*)));
                (*\hlplacehld{bitmap}{20}*).SetPixel((*\hlplacehld{x}{21}*), (*\hlplacehld{y}{22}*), Color.FromArgb((*\hlplacehld{newAlphaValue}{23}*), (*\hlplacehld{c}{24}*)));
            }
            else
                (*\hlplacehld{bitmap}{25}*).SetPixel((*\hlplacehld{x}{26}*), (*\hlplacehld{y}{27}*), (*\hlplacehld{c}{28}*));
        }
    }
}
...
\end{lstlisting}
\begin{description}
\item[\hlreflarge{14}] \id{alphaPercentage}: 52\%, \id{bitmap}: 32\%, \underline{\id{c}: 13\%}
\item[\hlreflarge{16}] \id{bitmap}: 67\%, \id{alphaPercentage}: 27\%, \underline{\id{c}: 4\%}
\item[\hlreflarge{17}] \underline{\id{alphaPercentage}: 85\%}, \id{c}: 6\%, \id{JPEQ\_QUALITY}: 3\%
\item[\hlreflarge{18}] \underline{\id{newAlphaValue}: 51\%}, \id{bitmap}: 24\%, \id{alphaPercentage}: 11\%
\item[\hlreflarge{19}] \underline{\id{newAlphaValue}: 86\%}, \id{y}: 4\%, \id{alphaPercentage}: 3\%
\item[\hlreflarge{20}] \underline{\id{bitmap}: 100\%}, \id{c}: 4e-3, \id{JPEG\_QUALITY}: 3e-4
\item[\hlreflarge{21}] \id{bitmap}: 98\%, \underline{\id{x}: 9e-2}, \id{c}: 8e-3
\item[\hlreflarge{22}] \id{c}: 50\%, \id{bitmap}: 49\%, \id{newAlphaValue}: 2e-3, \underline{\id{y}: 3e-8}
\item[\hlreflarge{23}] \id{alphaPercentage}: 42\%, \id{JPEG\_QUALITY}: 40\%, \id{bitmap}: 10\%, \underline{\id{newAlphaValue}: 3\%}
\item[\hlreflarge{24}] \id{newAlphaValue}: 60\%, \id{alphaPercentage}: 25\%, \underline{\id{c}: 5\%}
\item[\hlreflarge{25}] \underline{\id{bitmap}: 100\%}, \id{c}: 8e-4, \id{alphaPercentage}: 3e-4
\item[\hlreflarge{26}] \id{bitmap}: 88\%, \underline{\id{x}: 9\%}, \id{c}: 2\%
\item[\hlreflarge{27}] \id{c}: 79\%, \id{bitmap}: 18\%, \id{JPEG\_QUALITY}: 1\%, \underline{\id{y}: 3e-3}
\item[\hlreflarge{28}] \underline{\id{c}: 82\%}, \id{y}: 6\%, \id{x}: 4\%
\end{description}
\end{minipage}

\begin{minipage}{\textwidth}
\textbf{Sample 9}
\begin{lstlisting}[xleftmargin=0cm,frame=tlbr]
...
string s(*\hlref{1}*) = (string)(*\hlplacehld{Value}{2}*);
byte[] data(*\hlref{3}*);

Guid g(*\hlref{4}*);
if ((*\hlplacehld{s}{5}*).Length == 0)
{
    (*\hlplacehld{data}{6}*) = CollectionUtils.ArrayEmpty<byte>();
}
else if (ConvertUtils.TryConvertGuid((*\hlplacehld{s}{7}*), out (*\hlplacehld{g}{8}*)))
{
    (*\hlplacehld{data}{9}*) = (*\hlplacehld{g}{10}*).ToByteArray();
}
else
{
    (*\hlplacehld{data}{11}*) = Convert.FromBase64String((*\hlplacehld{s}{12}*));
}

SetToken(JsonToken.Bytes, data(*\hlref{13}*), false);
return data(*\hlref{15}*);
...
\end{lstlisting}
\begin{description}
\item[\hlreflarge{2}] \id{t}: 58\%, \underline{\id{Value}: 12\%}, \id{\_tokenType}: \%
\item[\hlreflarge{5}] \id{TokenType}: 44\%, \id{QuoteChar}: 43\%, \id{\_currentPosition}: 4\%, \underline{\id{s}: 9e-3}
\item[\hlreflarge{6}] \id{\_tokenType}: 74\%, \underline{\id{data}: 20\%}, \id{\_currentState}: 5\%
\item[\hlreflarge{7}] \id{QuoteChar}: 31\%, \id{ValueType}: 26\%, \id{Path}: 9\%, \underline{\id{s}: 3e-4}
\item[\hlreflarge{8}] \underline{\id{g}: 100\%}, \id{data}: 6e-5, \id{t}: 3e-5
\item[\hlreflarge{9}] \underline{\id{data}: 99\%}, \id{\_tokenType}: 5e-3, \id{ValueType}: 9e-4
\item[\hlreflarge{10}] \underline{\id{g}: 99\%}, \id{data}: 6e-3, \id{\_currentState}: 2e-3
\item[\hlreflarge{11}] \underline{\id{data}: 66\%}, \id{\_tokenType}: 31\%, \id{\_currentState}: 6e-3
\item[\hlreflarge{12}] \underline{\id{s}: 74\%}, \id{Value}: 20\%, \id{t}: 3\%
\end{description}
\end{minipage}

% ...

% ...

\section{Dataset}
\label{app:dataset}
The collected dataset and its characteristics are listed in \autoref{tbl:dataset}.

\newcommand{\dev}{$^{Dev}\xspace$}
\newcommand{\testonly}{$^\dag\xspace$}

\begin{table}[hp]
\caption{Projects in our dataset. Ordered alphabetically. kLOC measures the number of
non-empty lines of C\# code. Projects marked with \dev were used for
validation. Projects marked with \testonly were in the test-only dataset.
The rest of the projects were split into train-validation-test. The
dataset contains in total about 4,824kLOC.}\label{tbl:dataset}
\begin{tabular}{lrrrp{5.3cm}} \toprule
Name & Git SHA & kLOCs & plhldrs & Description \\ \midrule
Akka.NET& \id{9e76d8c} & 236 & 183k& Actor-based Concurrent \& Distributed Framework\\
AutoMapper& \id{6dd6adf} & 43& 21k& Object-to-Object Mapping Library \\
BenchmarkDotNet\testonly& \id{b4d68e9} & 23& 1k & Benchmarking Library\\
BotBuilder\testonly& \id{a6be5de} & 43 & 32k & SDK for Building Bots\\
choco& \id{73b7035} & 34 & 21k& Windows Package Manager\\
CommonMark.NET\dev& \id{e94800e} & 14& 7k & Markdown Parser\\
Dapper& \id{637158f} & 18& 1k& Object Mapper Library\\
EntityFramework& \id{fa0b7ec}  & 263& 184k& Object-Relational Mapper\\
Hangfire\testonly& \id{ffc4912} & 33& 32k& Background Job Processing Library\\
Nancy& \id{422f4b4} & 69 & 49k& HTTP Service Framework \\
Newtonsoft.Json& \id{744be1a} & 119& 70k& JSON Library\\
Ninject& \id{dbb159b} & 13& 3k& Code Injection Library\\
NLog& \id{3954157}  & 67 & 37k& Logging Library\\
OpenLiveWriter& \id{78d28eb} & 290 & 159k& Text Editing Application\\
Opserver& \id{c0b70cb} & 24& 16k& Monitoring System\\
OptiKey& \id{611b94a} & 27& 14k& Assistive On-Screen Keyboard \\
orleans& \id{eaba323} & 223& 133k& Distributed Virtual Actor Model \\
Polly& \id{b5446f6} & 30& 25k& Resilience \& Transient Fault Handling Library \\
ravendb\dev& \id{2258b2c} & 647& 343k & Document Database \\
RestSharp& \id{e7c65df} & 20& 20k& REST and HTTP API Client Library \\
roslyn& \id{d18aa15} & 1,997& 1,034k& Compiler \& Code Analysis\\
Rx.NET& \id{594d3ee} & 180 & 67k& Reactive Language Extensions \\
scriptcs\testonly&\id{ca9f4da} & 18& 14k& C\# Text Editor\\
ServiceStack& \id{b0aacff} & 205& 33k& Web Framework\\
ShareX& \id{52bcb52} & 125 & 91k& Sharing Application\\
SignalR& \id{fa88089} & 53& 35k& Push Notification Framework\\
Wox\testonly& \id{cdaf627} & 13& 7k& Application Launcher\\ \bottomrule
\end{tabular}
\end{table}

\end{document}